\pdfoutput=1
\documentclass[journal]{IEEEtran}

\usepackage{booktabs}

 \usepackage[pdftex]{graphicx}
\DeclareGraphicsExtensions{.pdf,.jpeg,.png}

\usepackage{amsmath}

\usepackage{algorithmic}

\usepackage{array}

\usepackage{url}
\usepackage{subfig}
\usepackage{tabularx}
\usepackage{adjustbox}
\usepackage{booktabs}
\usepackage[table,xcdraw]{xcolor}
\usepackage{multirow}
\usepackage{amsmath,bm}
\usepackage{hyphenat}

\usepackage{bm}
\usepackage{amssymb}
\usepackage{amsthm}
\usepackage{bbding}
\usepackage{pifont}
\usepackage{subfig}
\usepackage{float}
\usepackage{caption}

\newcommand{\thedataset}{AiRound}
\DeclareMathOperator*{\argmax}{arg\,max}
\DeclareMathOperator*{\argmin}{arg\,min}
\DeclareMathOperator*{\mode}{mode}

\begin{document}

\title{AiRound and CV-BrCT: Novel Multi-View\\Datasets for Scene Classification}

\author{Gabriel~Machado,~\IEEEmembership{}
        Edemir~Ferreira,~\IEEEmembership{}
        Keiller~Nogueira,~\IEEEmembership{}
        Hugo~Oliveira,~\IEEEmembership{}
        Pedro~Gama~\IEEEmembership{}
        and Jefersson~A.~dos~Santos~\IEEEmembership{}% <-this % stops a space
\thanks{Authors would like to thank NVIDIA for the donation of the GPUs that allowed the execution of all experiments in this paper. We also thank CAPES, CNPq (424700/2018-2), and FAPEMIG (APQ-00449-17) for the financial support provided for this research.}        

\thanks{Gabriel~Machado, Edemir~Ferreira, Hugo~Oliveira, Pedro~Gama and Jefersson~A.~dos~Santos are with the Department of Computer Science, Universidade Federal de Minas Gerais, Brazil; gabriel.lucas@dcc.ufmg.br}% <-this % stops a space

\thanks{Keiller~Nogueira is with Computing Science and Mathematics, University of Stirling, Stirling, FK9 4LA, Scotland, UK; kno@cs.stir.uk} % still not paying for the happy hour. fuckkkk haha
}
%\thanks{Manuscript received month day, 2005; revised August 26, 2015.}}

%\markboth{IEEE JOURNAL OF SELECTED TOPICS IN APPLIED EARTH OBSERVATIONS AND REMOTE SENSING,}%
%{Gabriel~Machado \MakeLowercase{\textit{et al.}}: AiRound and CV-BrCT: Novel Multi-View\\Datasets for Scene Classification}

\maketitle

\begin{abstract}
    It is undeniable that aerial/satellite images can provide useful information for a large variety of tasks. But, since these images are always looking from above, some applications can benefit from complementary information provided by other perspective views of the scene, such as ground-level images.
    Despite a large number of public repositories for both georeferenced photographs and aerial images, there is a lack of benchmark datasets that allow the development of approaches that exploit the benefits and complementarity of aerial/ground imagery.
    In this paper, we present two new publicly available datasets named \thedataset~and CV-BrCT.
    The first one contains triplets of images from the same geographic coordinate with different perspectives of view extracted from various places around the world. 
    Each triplet is composed of an aerial RGB image, a ground-level perspective image, and a Sentinel-2 sample.
    The second dataset contains pairs of aerial and street-level images extracted from southeast Brazil.
    We design an extensive set of experiments concerning multi-view scene classification, using early and late fusion. Such experiments were conducted to show that image classification can be enhanced using multi-view data.
  %\textbf{The dataset and source codes for the experiments will be made available upon the acceptance of the paper.}
\end{abstract}

\begin{IEEEkeywords}
Remote Sensing, Deep Learning, Data Fusion, Multi-Modal Machine Learning, Dataset, Feature Fusion.
\end{IEEEkeywords}

%\ifCLASSOPTIONpeerreview
%\begin{center} \bfseries EDICS Category: 3-BBND \end{center}
%\fi
%\IEEEpeerreviewmaketitle

\section{Introduction} \label{sec:introduction}

% \todo{REVISORES:
% - The text refers to the two types of imagery without ground-based acquisition and indicates that "both are aerial". This is incorrect; one is aerial and one is satellite-based.
% - In fact, the acquired multi-spectral Sentinel-2 data is satellite imagery and not aerial imagery.
% }

%\todo{ARRUMAR INTRO DE ACORDO COM O NOVO OBJETIVO DO PAPER.}

Satellite images become more accessible to civilian applications each year. 
New technologies are enabling the wide usage of better and cheaper images in comparison with the past few decades.
Nowadays, it is also possible to access many free remote sensing image repositories with a variety of spatial, spectral and temporal resolutions~\cite{gorelick2017google}.
Images with aerial perspective give us a unique view of the world, allowing the capture of relevant information (not provided by any other type of image) that may assist in several applications, such as automatic geographic mapping and urban planning.
%In addition to satellite-born sensors, the popularization of drones is contributing to the acquisition of huge volumes of aerial images in a more feasible way than in the past. 
%These image sets tend to cover different geographical scales in comparison with satellite images and, therefore, it makes sense to fuse or combine them to better address some applications~\cite{shakhatreh2019unmanned}.

Despite the clear benefits of 
%acquiring and accessing large sets of 
optical aerial imagery, the fact that they are always looking from above may make their use limited.
Precisely, the presence of vegetation cover, clouds or simply the need of more detailed on-the-ground information can decrease the effectiveness of such images in some applications. %\ede{SAR é aereo e nao tem problemas com nuvens. No caso é uma mistura de problemas, aonde a pespectiva aerea dificulta a classificação mesmo com a possibilidade de extrair o contexto}
%In these cases, ground images are more recommended, given that they provide high-detailed information about the scene without suffering from the aforementioned issues.
%Fortunately, the popularization and cheapness of cameras on mobile devices associated with the expansion of the internet and data sharing platforms have allowed us to create rich repositories of accessible and georeferenced ground images.
%In this sense, some platforms that provide structured data such as Google Street View, OpenStreetCam, Mapillary further facilitate the development of automated image-based geolocation~\cite{anguelov2010google,leon2019value}.
%In some applications, using only aerial or ground images may not be enough, given that they may not provide sufficient information for solving the task.
% Some applications may require aerial and ground information
In multi-view scenarios, it would be crucial to combine the complementary information of aerial and ground images in order to efficiently tackle a problem.
Such combination of multiple sources images can benefit many applications in different fields, like 3D human pose estimation~\cite{h_early2}, places geo-localization~\cite{hu2018cvm}, and urban land use~\cite{srivastava2019understanding}.
%Because of this and supported by the fact that aerial and ground images have distinct properties and characteristics (including different notion of perspective and context), it makes sense to fuse or combine them to better address some applications~\cite{shakhatreh2019unmanned}.
%In the recent computer vision literature, it is possible to find many works that exploit ground/aerial images to obtain effective solutions~\cite{srivastava2019understanding,hoffmann2019model,lefevre2017toward,obj_detection}.
Motivated by these benefits, several approaches~\cite{hoffmann2019model,ghouaiel2016coupling,wegner2016cataloging,cao2018integrating,liu2019lending} have been proposed to exploit multi-view datasets to face distinct tasks.
%Furthermore, with the recent success of deep learning in many computer vision tasks, most of these existing works exploit Convolutional Neural Networks (ConvNets) to learn representations for these datasets.
%However, there is one huge challenge in exploring such datasets with deep learning: these techniques require a large amount of annotated data in order to converge its model~\cite{goodfellow2016deep}.
Although important, it is not easy to find multi-view datasets for image-related tasks, given the difficulty in creating and labeling such data.
% general purpose
In fact, as far as we know, there is no other publicly available multi-view (aerial and ground) dataset for image classification tasks in the literature.

%The great advances observed in the field of computer vision from the use of deep learning-based approaches poses us several challenges and research opportunities regarding aerial/ground scene classification in which we can list: (i) new multi-view fusion strategies; (ii) new cross-view transfer learning strategies; (iii) new strategies in witch given a query image retrieves its pair from another domain; and (iv) aerial to ground domain adaptation (and vice-versa).

%Supported by all this, 
In this paper, we present two novel multi-view images datasets. 
The main purpose of creating these datasets is to make them publicly available so that the scientific community can carry out image classification experiments in multi-view scenarios.
One of the datasets is composed of $1,165$ triplets of images, each one of those consisting of a ground scene, a high-resolution aerial image, and a multi-spectral aerial data. The images are unevenly divided into 11 classes, including airport, bridge, church, forest, lake, river, skyscraper, stadium, statue, tower, and urban park.
An interesting property of our dataset is that it was designed to contain a high inter-class variety, so it was selected places from all around the world to compose the samples.
The other dataset is composed of $24k$ pairs of images, each one containing a street-level scene and a high-resolution aerial image. Those samples are labeled in $8$ different classes, which includes apartment, hospital, house, industrial, parking lot, religious, school, store, and vacant lot.
Both datasets were evaluated for image classification, using early and late fusion strategies.
%It is important to mention that both datasets are proposed to be used in several image classification applications, including scene classification~\cite{srivastava2019understanding,hoffmann2019model}, cross-view matching
%~\cite{cvusa,liu2019lending} and multi-modal machine learning tasks ~\cite{sun2015unsupervised,hoffmann2019model,srivastava2019understanding}.
%It is important to mention that besides both datasets are proposed to be mainly used in image classification tasks, they can be also used in a lot of other different applications.
It is important to emphasize that, although we assessed the performance of both dataset for image classification, they were proposed to be used in distinct image-related tasks, varying from image classification to cross-view matching and multi modal learning.
%Those applications involve knowledge of multiple fields in machine learning, including deep feature fusion~\cite{srivastava2019understanding,hoffmann2019model}, cross-view matching~\cite{cvusa,liu2019lending}, aerial-ground domain adaptation~\cite{sun2015unsupervised,hou2015unsupervised}, and other multi-modal machine learning tasks.
%In this work, we evaluated several image classification and feature fusion methods using the proposed datasets.
%, including image classification and feature fusion.
%\todo{Jefersson, justificar melhor o pq avaliamos somente essas duas tarefas}

%\todo{Objectives: 1) define a concise benchmark evaluation platform for cross-view image classification and retrieval; 2) provide pretrained models for similar cross-view applications.}

%In summary, these are the contributions of this work:
In summary, the contributions of this work are:
(i) two novel multi-view scene classification datasets, named~\thedataset~and CV-BrCT,
(ii) a full evaluation of the proposed datasets in image classification tasks using several deep learning state-of-the-art methods and late fusion techniques,
(iii) a novel methodology of performing early feature fusion using state-of-the-art deep architectures as a base.
%(iv) a full evaluation of experiments performing some of the most famous decision-based (late fusion) techniques.

%\todo{Outline contributions: 1) dataset; 2) benchmark for cross-view classification and retrieval; 3) evaluation of different late fusion methods and DNN architectures; 4) evaluate the effect of several UDA techniques in order to merge the information of all views.}

The remainder of this paper is organized as follows. 
Section~\ref{sec:related_work} presents related work.
The proposed datasets are presented in Section~\ref{subsec:dataset}, while Section~\ref{sec:methods} introduces the methods and tasks evaluated using these datasets.
The experimental setup is introduced in Section~\ref{sec:experimental_setup} while Section~\ref{sec:results} presents the obtained results. 
Finally, Section~\ref{sec:conclusion} concludes the paper.

%\todo{definir o que eh multi-view (usei esse termo)}
%\todo{novas contribuiçoes: 2 primeiros datasets (publicos) para aerial/ground scene classification. Experimentos envolvendo fusão early e late. Metodologia para arquiteturas early. Primeiro paper a avaliar impacto de fusão early em aerial/ground scene classification.}
\section{Related Work} \label{sec:related_work}

% \todo{Nessa secao, vc tem que descrever os datasets existentes e explicar o pq estamos propondo outros datasets. O que eles tem a mais que os outros nao tem?
% Uma sugestão nessa seção eh incluir exemplos visuais dos outros datasets. Mas acho que isso so vale a pena se der pra mostrar/analisar algumas diferencas entre os datasets existentes e propostos, mostrando que os datasets propostos aqui tem coisas que os outros nao tem.}

%Since the emergence of deep learning, the importance of existing distinct datasets for different tasks have been increasing a lot.
%Considering the typically state-of-the-art results that deep learning achieves for almost all kinds of image data, and that the techniques, which are based on its concept, usually demands a high amount of data, a lot of datasets emerged in the literature. 
%Naturally, some of those datasets contains remote sensing data, and those have been used for several tasks, such as image classification, change detection and semantic segmentation.

Considering recent advances in satellite data acquisition and cloud computing, access to high-resolution satellite images and other types of data was facilitated. 
Despite the great advantages that aerial images provide, some applications demand information that an aerial perspective may lack. 
In these cases, an alternative solution is to use complementary perspectives of the same view, i.e., ground-level view, to better seek these information~\cite{ex1,ex2,ex3,cv-6}. 
%In fact, the possibility of exploring the completarity 
%Even with the accessibility to data and the existence of techniques to handle them, in some cases, the use of remote sensing images alone is not enough~\cite{ex1,ex2,ex3,cv-6}. %Some tasks, like image geo-localization or cross-view image retrieval, can demand information that RS images alone do not contain.
Due to the high demand for images to be used by those kinds of tasks, a lot of multi-view datasets were proposed in the literature.
In Table~\ref{tab:datasets}, we summarized some of the most similar datasets compared to the novel ones proposed for this work.
%Some of those datasets, that are most similar to the ones proposed for this paper are summarized in Table~\ref{tab:datasets}.

%\todo{COLOCAR CITACAO PARA AS APIs.}

% Firstly, looking at multi-view datasets proposed for retrieval tasks (cross-view matching), the first one was named CV-USA and it was published by Workman \textit{et al}.~\cite{cvusa}. 
% This dataset contains millions of pairs of aerial and ground images, that were taken from across the United States. 
% Relating to its data collection, the aerial images were collected using Bing Maps API, and the ground images used Flickr and Google Street View APIs. 
% Finally, it is important to mention that besides this dataset have millions of samples, most of the works use a subset of it, which have around $44k$ images.
% The other image retrieval dataset with multi-view data was named CV-ACT and it was proposed by Liu \textit{et al.}~\cite{liu2019lending}. 
% This dataset contains approximately $128k$ images, that were taken covering a dense area of the city Canberra. The images were collected using Google Maps and Google Street View APIs.
The CV-USA~\cite{cvusa} and CV-ACT~\cite{liu2019lending} datasets were proposed specifically for retrieval tasks, i.e., cross-view matching.
The first one contains millions of pairs of aerial and ground images, that were taken from across the United States. 
Relating to its data collection, the aerial images were collected using Bing Maps API (BM), and the ground images used Flickr and Google Street View API (GSV). Another important aspect to mention is that even if CV-USA has millions of samples, most of the works use a subset of it, which have around $44k$ images.
Relating to the CV-ACT dataset, it contains approximately $128k$ images, that were taken covering a dense area of the city Canberra. All its images were collected using Google Maps (GM) and GSV APIs.
Similarly, Cities~\cite{lin2015learning} and Urban Environments~\cite{tian2017cross} datasets were designed to tackle cross-view matching problem, but both of them were not publicly released. The first one used Google APIs to collect pairs of images from cities around the world. The latter one collected pairs of images from Pittsburg, Orlando and Manhattan using GSV and BM APIs.

% Secondly, focusing on multi-view instance segmentation and object detection datasets presented in Table~\ref{tab:datasets}, two of them were highlighted.
% The first one, named Pasadena Urban Trees was proposed by Wegner \textit{et al.} \cite{wegner2016cataloging}. 
% This dataset was designed for object detection and used OpenStreetMap (OSM) annotations of trees in the city of Pasadena. It contains $18$ different species of trees, which ground samples were collected using Google Street View API, and the aerial ones used Google Maps API.
% The second dataset, named Brooklyn and Queens was proposed for instance segmentation by Workman \textit{et al.}~\cite{workman2017unified}. This dataset contains approximately $213k$ images of $206$ different types of buildings, covering areas from the two boroughs of New York City, that named the dataset. 
% All the images from this dataset were collected using Bing Maps and Google Street View APIs.
The Pasadena Urban Trees~\cite{wegner2016cataloging} was designed for object detection. This dataset used OpenStreetMap's (OSM) bounding box annotations of trees in the city of Pasadena. 
It contains $18$ different species of trees, which ground samples were collected using GSV API, and the aerial ones used Google Maps (GM) API.

Another multi-view dataset was named Brooklyn and Queens~\cite{workman2017unified}, and it was proposed for instance segmentation. 
It contains approximately $213k$ images of $206$ different types of buildings, covering areas from the two boroughs of New York City. 
All the images from this dataset were collected using BM and GSV APIs, and it was used OSM to define the labels of all samples.

%Multiple remote sensing classification datasets were proposed over the year  WHU20~\cite{hu2015benchmark}, 
%EuroSAT~\cite{helber2019eurosat},
%BrazilDAM~\cite{ferreira2020brazildam}

% Lastly, looking at multi-view image classification [...]
% - WHU20
% - EuroSAT
% - Buildings
% - Ile-de-France land use
% - BrazilDAM

% Thirdly, relating to multi-view scene classification datasets, we will highlight $2$ datasets taken from Table \ref{tab:datasets}, which are the ones that are most similar to the datasets proposed on this paper. 
% Those datasets were proposed by Hoffmann \textit{et al.}~\cite{hoffmann2019model} and Srivastava \textit{et al.}~\cite{srivastava2019understanding}, and they were named Buildings and Île-de-France land use, respectively.
% The first dataset contains $56,259$ paired aerial/street-level images of $4$ different types of buildings, covering 49 different states across the US, Washington DC, and Puerto Rico. 
% All the building labels from this dataset were taken using annotations contained in OSM, and relating to the data collection, for aerial images were collected images from 3 different zoom levels of each building using Google Maps API, while for street-level images Google Street-View API was used.
% Relating to the Île-de-France land use dataset, it contains approximately $25k$ pairs of aerial/ground images of 16 different land use classes, covering the metropolitan region of Paris and some nearby suburbs. 
% This dataset also uses OSM to collect its labels, and the same APIs of Buildings dataset to collect images.

Relating to the Buildings~\cite{hoffmann2019model} and Île-de-France land use~\cite{srivastava2019understanding} datasets, both were designed for multi-view scene classification. The first dataset contains $56,259$ paired aerial/street-level images of $4$ different types of buildings, covering Washington DC. Puerto Rico, and $49$ different states across the US. 
Relating to the first dataset, all of its building labels were defined using annotations contained in OSM. The data collection was made using two different APIs, being those, GM API for aerial samples, and GSV API for the ground perspective ones. 
The Île-de-France land use dataset contains approximately $25k$ pairs of aerial/ground images of 16 different land use classes, covering the metropolitan region of Paris and some nearby suburbs. 
This dataset also uses OSM to collect its labels, and the same APIs of the Buildings dataset to collect the samples.

%After the success of deep learning for remote sensing applied to computer vision tasks, other datasets, designed for other kinds of tasks, appeared in the literature.

%Given the fact that those tasks generally are increasingly complex, to properly solve them naturally require a higher amount of information. 
%One of the ways to tackle this issue is to use complementary data, which generally contains information that a remote sensing image alone can not capture. 
%In fact, there are datasets and studies that encapsulate data from other modalities and compare the impact of their use. 
%So, for this related work, we selected some of those multi-view datasets, that are most similar to the ones proposed for this paper and summarized them in Table \ref{tab:datasets}.

\begin{table*}[!ht]
\centering
\begin{adjustbox}{max width=\textwidth}
\begin{tabular}{@{}cccccccccc@{}}
\toprule
\multirow{2}{*}{\textbf{Dataset}} & \multicolumn{3}{c}{\textbf{Image Type}} & \multirow{2}{*}{\textbf{\begin{tabular}[c]{@{}c@{}}Publicly \\ Available\end{tabular}}} & \multirow{2}{*}{\textbf{\begin{tabular}[c]{@{}c@{}}Paired \\Aerial/Ground\\ Images\end{tabular}}} & \multirow{2}{*}{\textbf{\begin{tabular}[c]{@{}c@{}}Total of \\ Samples\end{tabular}}} & \multirow{2}{*}{\textbf{\begin{tabular}[c]{@{}c@{}}Number of \\ Classes\end{tabular}}} & \multirow{2}{*}{\textbf{Task}} & \multirow{2}{*}{\textbf{Year}} \\ \cmidrule(lr){2-4}
 & \multicolumn{1}{l}{\textbf{Aerial RGB}} & \multicolumn{1}{l}{\textbf{Ground}} & \multicolumn{1}{l}{\textbf{Multispectral}} &  &  &  &  &  &  \\ \midrule
CV-USA~\cite{cvusa} & \ding{51} & \ding{51} & \ding{55} & \ding{51} & \ding{51} & $\sim44k$ & - & Cross-View Matching & 2015 \\
%WHU20~\cite{hu2015benchmark} & \ding{51} & \ding{55} & \ding{51} & \ding{51} & \ding{55} & 5k & 20 & Classification & 2015 \\
Cities~\cite{lin2015learning} & \ding{51} & \ding{51} & \ding{55} & \ding{55} & \ding{51} & $\sim156k$ & - & Cross-View Matching & 2015 \\
Pasadena Urban Trees~\cite{wegner2016cataloging} & \ding{51} & \ding{51} & \ding{55} & \ding{51} & \ding{51} & $\sim100k$ & 18 & Object Detection & 2016 \\
Brooklyn and Queens~\cite{workman2017unified} & \ding{51} & \ding{51} & \ding{55} & \ding{51} & \ding{51} & $\sim213k$ & - & Instance Segmentation & 2017 \\
Urban Environments~\cite{tian2017cross} & \ding{51} & \ding{51} & \ding{55} & \ding{55} & \ding{51} & $\sim18k$ & - & Cross-View Matching & 2017 \\
%NWPU-RESISC45~\cite{cheng2017remote} & \ding{51} & \ding{55} & \ding{55} & \ding{51} & \ding{55} & 31.5k & 45 & Classification & 2017 \\
%AID~\cite{xia2017aid} & \ding{51} & \ding{55} & \ding{55} & \ding{51} & \ding{55} & 10k & 30 & Classification & 2017 \\
%EuroSAT~\cite{helber2019eurosat} & \ding{51} & \ding{55} & \ding{51} & \ding{51} & \ding{55} & 27k & 10 & Classification & 2019 \\
CV-ACT~\cite{liu2019lending} & \ding{51} & \ding{51} & \ding{55} & \ding{51} & \ding{51} & $\sim128k$ & - & Cross-View Matching & 2019 \\
Buildings~\cite{hoffmann2019model} & \ding{51} & \ding{51} & \ding{55} & \ding{55} & \ding{51} & $\sim261k$ & 4 & Classification & 2019 \\
Île-de-France land use~\cite{srivastava2019understanding} & \ding{51} & \ding{51} & \ding{55} & \ding{55} & \ding{51} & $\sim$50k & 16 & Classification & 2019 \\
%BrazilDAM~\cite{ferreira2020brazildam} & \ding{51} & \ding{55} & \ding{51} & \ding{51} & \ding{55} & 1.9k & 16 & Classification & 2020 \\
\hline
\textbf{AiRound (ours)} & \ding{51} & \ding{51} & \ding{51} & \ding{51} & \ding{51} & $\sim$3.5k & 11 & Classification & 2020 \\
\textbf{CV-BrCT (ours)} & \ding{51} & \ding{51} & \ding{55} & \ding{51} & \ding{51} & $\sim$48k & 9 & Classification & 2020 \\ \bottomrule
\end{tabular}
\end{adjustbox}
 \caption{Properties of other datasets found in the literature that are similar to \thedataset~and CV-BrCT.}
 \label{tab:datasets}
\end{table*}

Differentiating our datasets from the ones in Table~\ref{tab:datasets}, some of the existing datasets were designed in a way that each image pair can be seen as a class. 
Such datasets do not contain groups of classes that share the same label, which ends up making its use for image classification unenviable. Other datasets are quite different from the ones proposed here, given that the main task for which they were proposed is different. That difference mainly comes, because those problems require different types of labels as inputs and also generates distinct outputs (bounding boxes and segmentation).
Lastly, relating to multi-view image classification datasets, two datasets~\cite{hoffmann2019model,srivastava2019understanding} are quite similar to both datasets proposed here. 
However, neither of these existing datasets are publicly available, while ours will be.

\section{Proposed Datasets}
\label{subsec:dataset}

%\todo{Essa secao agora pode ser extensiva, pois nao temos limite de pagina, temos? Detalhe bastante o dataset, analise a distribuicao das classes, as caracteristicas, etc.
%REVISORES: 
%- Please describe the used Sentinel images in more detail: Are composites used? Which acqusition dates are used? Which pre-processing is used or which products are used? The same applies to the aerial photographs. Necessary key information should be mentioned.
%}
In this work, we proposed two novel multi-view datasets. 
%Like was said before, as far as we know there is not any publicly available dataset for multi-view scene classification, so we proposed two new datasets for this work. 
\textbf{It is important to mention that both datasets are publicly available for research purposes at the project's website\footnotemark\footnotetext{\url{http://www.patreo.dcc.ufmg.br/multi-view-datasets/}}.} 
Since both datasets were designed using a different methodology, in the following sections we will describe the relevant characteristics of each one and the methodology used to collect the samples.

\subsection{The \thedataset~Dataset} \label{subsec:airound_dataset}

%Most of the aerial/ground datasets found in the literature are intended for image retrieval.
%For instance, the CVUSA~\cite{cvusa} and CVACT~\cite{liu2019lending} datasets pair aerial/ground images from streets of multiple cities around the world.
%Furthermore, one can also easily find aerial/ground datasets for tasks such as object detection~\cite{wegner2016cataloging}, semantic segmentation~\cite{cao2018integrating} and scene classification~\cite{hoffmann2019model,srivastava2019understanding}.
% To the best of our knowledge
%However, as far as we know, the only aerial/ground dataset~\cite{hoffmann2019model} proposed for general image classification applications is not publicly available.
%Thus, we propose a new publicly available multi-purpose aerial/ground scene dataset, \textbf{henceforth referred to as \thedataset}.

% Most of the cross-view scene datasets found in the literature are intended for image retrieval evaluation, where images that were taken are from streets of different cities around the world (CVUSA \cite{cvusa} and CVACT \cite{crossview4}). We also found cross-view datasets for other tasks, like object detection dataset ~\cite{wegner2016cataloging}, semantic segmentation \cite{cao2018integrating} and scene classification \cite{hoffmann2019model}.
% Since the only cross-view scene classification dataset that we found in the literature is not publicly available \cite{hoffmann2019model}, a new dataset was proposed for this work.

\begin{figure}[!ht]
    \centering
    \includegraphics[width=0.95\columnwidth]{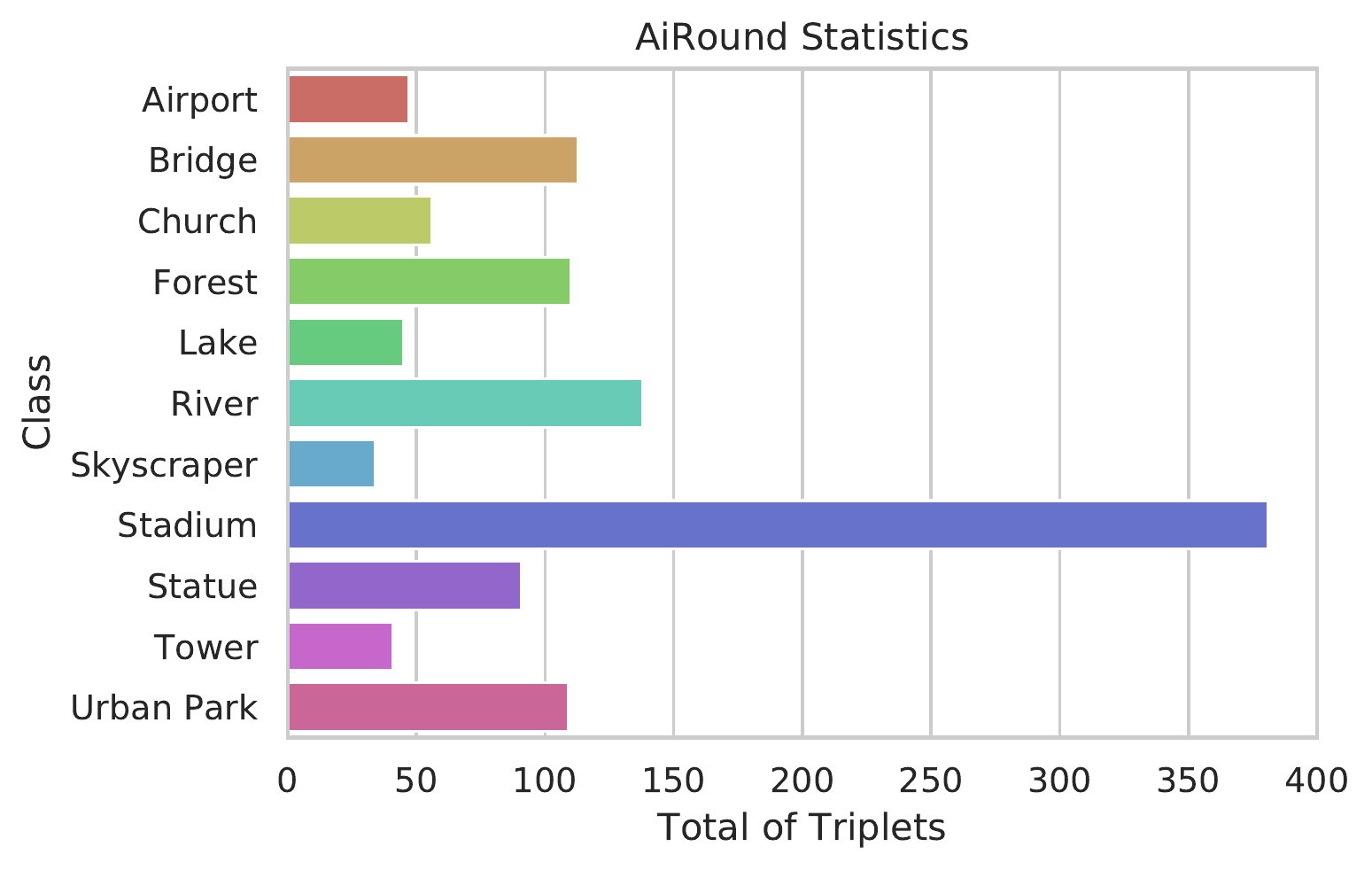}
    \caption{Class distribution of the proposed \thedataset~dataset. Note that each image is represented by a triplet of ground, aerial, and multispectral data.}
    \label{fig:dataset_hist}
\end{figure}

\begin{figure*}[!ht]
    \centering
    \includegraphics[width=0.95\textwidth]{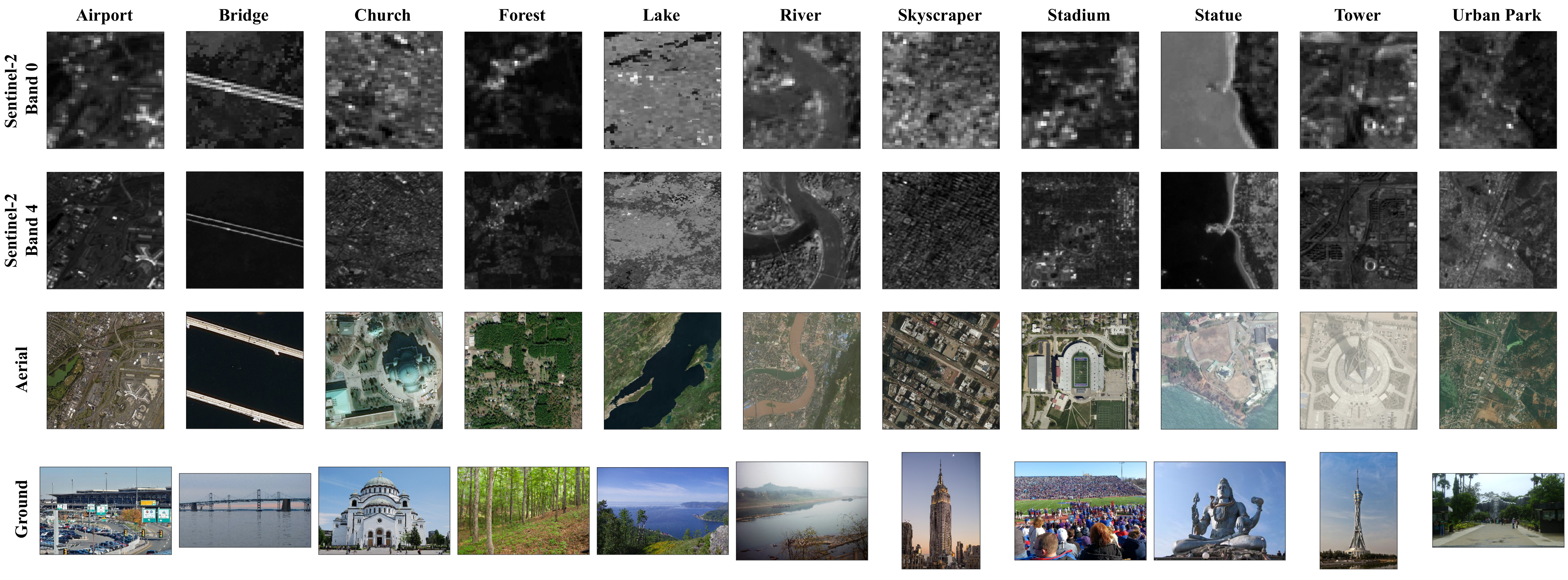}
    \caption{Examples of instances taken from \thedataset. The two top rows show channels of a sentinel-2 sample, while the third and fourth rows show the high resolution aerial perspective image and the ground view image, respectively.}
    \label{fig:dataset}
\end{figure*}

The first dataset is named \thedataset, and is composed of $3,495$ images distributed among 11 classes, including: airport, bridge, church, forest, lake, river, skyscraper, stadium, statue, tower, and urban park.
Each sample is composed by a triplet, that contains images in 3 distinct points of view: (i) a ground perspective image; (ii) a high resolution RGB aerial image; and (iii) a multi-spectral image taken from the Sentinel-2 satellite.
All images are paired and were manually checked to guarantee their correctness. %, allowing for verified quantitative comparisons between different classification and retrieval methods. 
The distribution of samples from \thedataset~can be checked in Figure~\ref{fig:dataset_hist} and examples of instances can be seen in Figure~\ref{fig:dataset}. 
%\textbf{Finally, it is important to mention that \thedataset~and all its gathering and benchmark source codes are publicly available for research purposes at the project's website\footnotemark\footnotetext{\url{omitted for blind review}}.}

The data collected in this dataset are directly linked to real places around the world.
To download the samples, two types of metadata were required: (i) the name of the place; and (ii) its correspondent geographical coordinates.
%Those information were collected using crawlers in specific Wikipedia webpages \ede{ Quais paginas? Como foi escolhido? Essa seleção tem que ser justificada a high variance}.
%Those information were collected using web crawlers in diversified lists of Wikipedia webpages. As instance, a list of tallest buildings [ref] was used to ensure that samples have been extracted from different parts of the world.
Those information were collected using web crawlers in diversified lists of Wikipedia web pages. As instance, a list of tallest buildings\footnotemark\footnotetext{\url{https://en.wikipedia.org/wiki/List_of_tallest_buildings}} was used to ensure that samples from building class have been extracted from different parts of the world. For more details about the web pages used and all the metadata required to create \thedataset~dataset, we recommend checking the project's websitee\footnotemark\footnotetext{\url{http://www.patreo.dcc.ufmg.br/multi-view-datasets/}}.

%Finally, in order to download the images, specific APIs were used to collect the samples for each modality. %those APIs are properly described and the methodology used is detailed in the following paragraphs.  
%The details of which API was used and the resolution of the images from each view can be seen in Table~\ref{tab:api}.

Given the metadata, the RGB aerial images were collected using Bing Maps API\footnote{\url{https://docs.microsoft.com/en-us/bingmaps/}}. %In order to download each one of the images, the places' geographical coordinates were used as input in the query.
The zoom level was empirically selected in order to adapt a proper vision for the samples of each class. 
Since there is a huge difference in areas occupied for some classes (river and skyscraper, for instance), this zoom level ended with large variance, specifically between $5$ and $19$, which corresponds to a spatial resolution that varies from 4891.97 to 0.30 meters per pixel.
Finally, it is important to mention that all aerial images downloaded have a image size of $500 \times 500$ pixels.

In order to collect the ground level samples, it was checked if the correspondent class exists in the Google Places' database. 
If the sample class exists, a query was built using this place's geographical coordinates as input. 
The outputs returned by this API were all manually checked, and if they do not correspond to the class, another protocol was performed. 
The second protocol was used for cases that the class did not exist in Google Places' database or the image retrieved did not correspond to the query requested. 
This protocol consists of crawling the top 5 images from Google Images using, as query, the place's name. Finally, it was manually selected to represent each sample on \thedataset~the best instance between the 5 images downloaded.
It should be pointed out, that the resolution of each sample is not standardized because the methodology employed does not allow the selection of a specific resolution.
%does not have a pattern, and that is because of the methodology that was used to download them does not allow to select a specific resolution.
Lastly, it is important to mention that, besides most of the papers use Google Street-View API to download street-level data, we could not use such API for this dataset, because it is not capable of downloading images for some classes, like river or lake, for instance.

Finally, concerning to the Sentinel-2 images acquisition, we followed exactly the same protocol that was proposed by Ferreira \textit{et al.}~\cite{ferreira2020brazildam}. In this protocol, Google Earth Engine~\cite{gorelick2017google} was used to download the data using the place's geographical coordinates.
After careful analysis, we decided to resize all images to 300x300 pixels, a resolution that could cover all the classes' areas.

\subsection{CV-BrCT} \label{subsec:CV-BrCT}

%\todo{Pedro, montar um gráfico com as quantidades de img de cada classe.}
The CV-BrCT dataset, which stands for Cross-View Brazilian Construction Type, comprises of approximate $24k$ pairs of images split into 9 urban classes. The pairs are composed of images from two different views: an aerial view, and a frontal view of a location. This dataset is focused on the urban environment and the 9 classes are:

\begin{itemize}
\item Apartment: Buildings with at least two stories primarily for residential use.
\item Hospital: Health-related constructions, primarily hospitals but can include small particular clinics.
\item House: a single-family residence.
\item Industrial: Manufactured related buildings. Includes large storage constructions.
\item Parking Lot: Includes, both open and indoors parking lots.
\item Religious: Religious buildings; this includes catholic churches, protestant churches.
\item School: Any school construction. From elementary schools to high school. 
\item Store: Any commercial or service related building.
\item Vacant Lot: Demarked areas without construction. It can include abandoned open areas.
\end{itemize}

\begin{figure}[!ht]
    \centering
    \includegraphics[width=0.95\columnwidth]{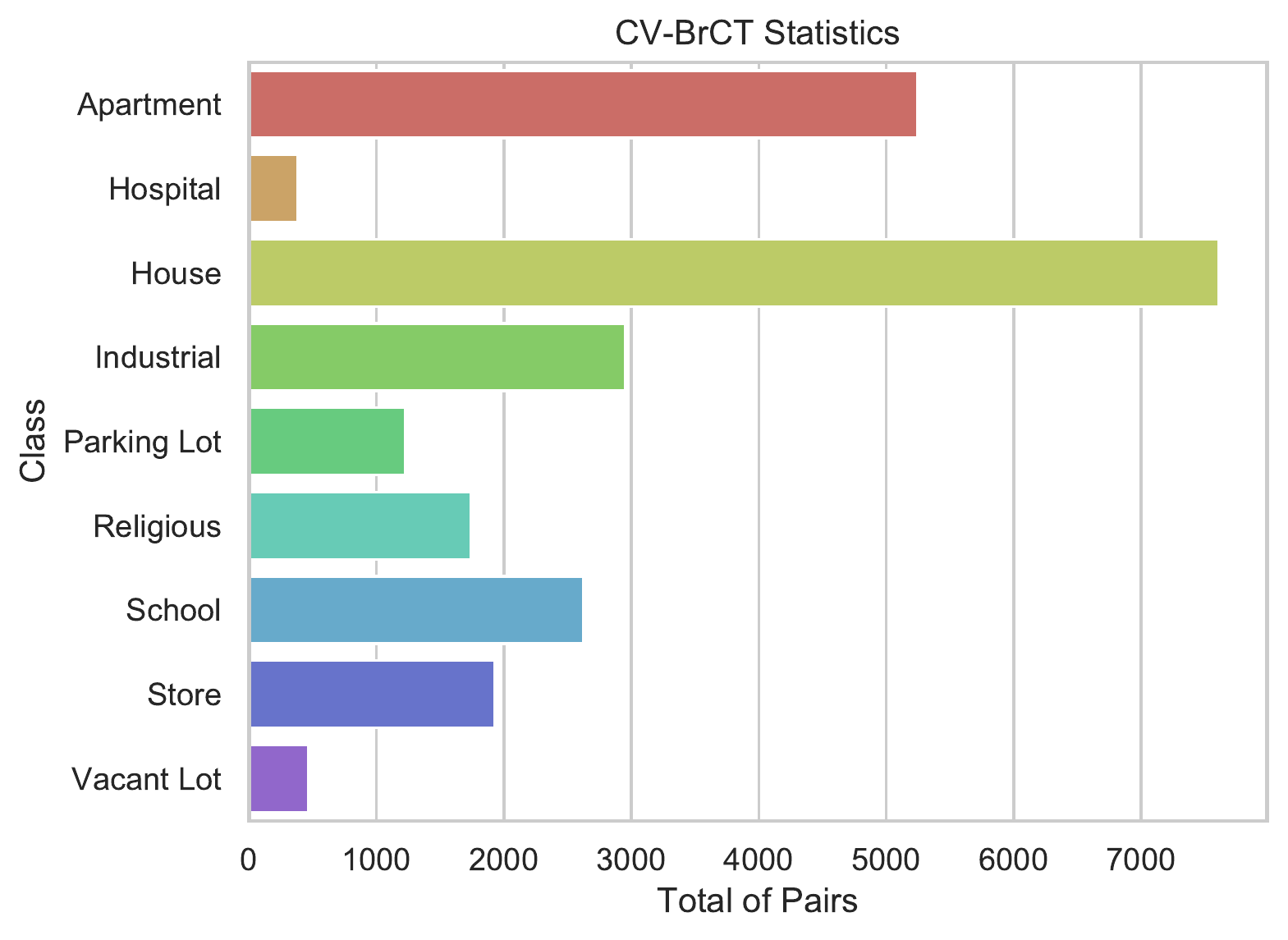}
    \caption{Class distribution of the proposed CV-BrCT dataset.}
    \label{fig:cv-brct-stats}
\end{figure}

\begin{figure*}[!ht]
    \centering
    \includegraphics[width=0.95\textwidth]{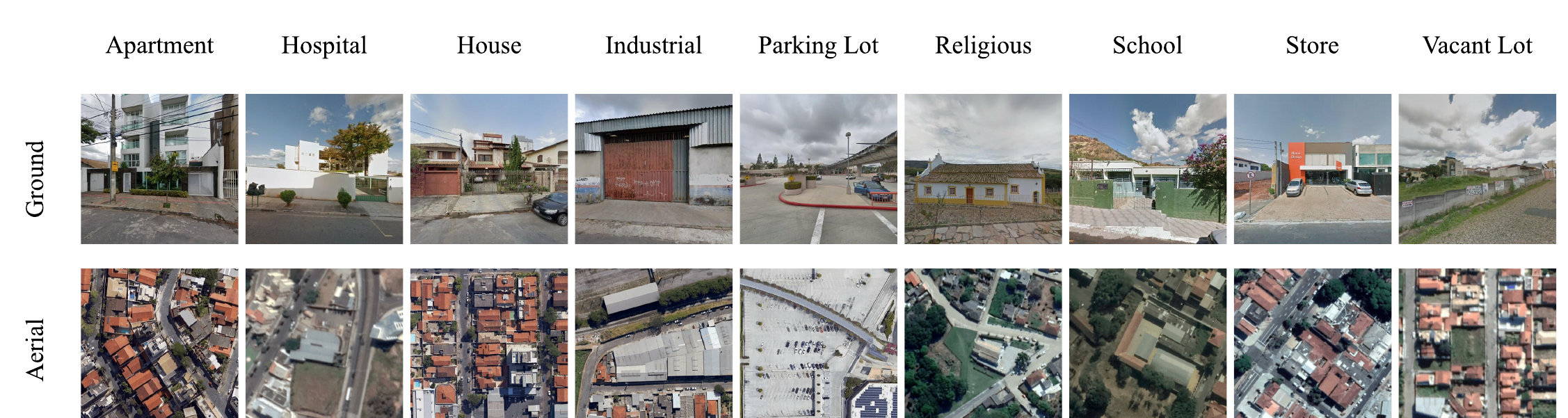}
    \caption{Examples of instances taken from CV-BrCT.}
    \label{fig:ex_cvbrct}
\end{figure*}

Examples can be seen in Figure \ref{fig:ex_cvbrct}, whereas the class distribution is presented in Figure \ref{fig:cv-brct-stats}. Regarding the images, all of them are 500$\times$500 RGB images. As implied by the name, this dataset contains only Brazilian locations. These are mainly in the Southeast region of Brazil, specifically the states of Minas Gerais and Sao Paulo, with some classes adding locations from states from other regions, i.e Goias in the Center-West region.

The data was collected following a simple protocol. For all the classes, a list of geographical coordinates was generated where each coordinate represents a location. Except for the Vacant Lot class, that was manually annotated, these lists were obtained from the publicly available data of the OpenStreetMap\footnote{\url{www.openstreetmap.org/}}, a community-based project were users annotate aerial images to create maps, and collected using the Overpass API\footnote{\url{https://overpass-turbo.eu/}}. As the data is provided by users, not necessarily specialists, they can contain they can contain poorly annotated samples which can lead to outliers in the dataset. The lists are then fed to scripts that utilize the Google StaticMap API\footnote{\url{https://developers.google.com/maps/documentation/maps-static/intro}}, to collect the aerial images, and the Google StreetView API\footnote{\url{https://developers.google.com/maps/documentation/streetview/intro}}, to collect the frontal images. 
With the exception of the zoom parameter, which was set to 19 empirically, the default values of the the Google APIs were used for the aerial images. 
%The parameters used in the Google APIs were mostly set to default values, but the zoom level hyperparameter (19 were used) for the aerial images were empirically selected to provide satisfactory images for the majority of the locations. \ede{With the exception of the zoom parameter, which was set to 19 empirically, the default values of the the Google APIs were used.} 
As we gathered a large collection of locations, we ignored points points where the StreetView API could not retrieve an image.

As a final step, an additional removal of outliers was applied after all the images were collected. This final filter consisted of firstly obtain a feature vector of the frontal images.  These feature vector were produced by a ResNet pre-trained on the ImageNet dataset, collected from the final layer of the architecture. Then, for each class, a k-means++~\cite{arthur2006k} clusterization was applied using these feature vectors. With the clusters, the distance of each data point, within a class, was calculated to its closest centroid as well as the mean and standard deviation distance of each cluster. Points that were more than 3 s.d. away from a centroid cluster were removed from the dataset. 

Even with these removal operations, by the nature of the data collection and the simplicity filters applied, it is possible that noise is present in the dataset. However, we assume that the noise is minimal after all the process.

\section{Benchmarched Methods} \label{sec:methods}

%As aforementioned, several methods were tested for multi-view scene classification and aerial/ground features fusion.
%Such techniques are further described in the next sections.
This Section presents the evaluated methods. Different approaches were tested for multi-view scene classification. In order to better assess the improvement provided by combining distinct sources of data, we first evaluate the use of distinct networks for single-view data. Then, we evaluate the use of early and late fusion to perform multi-view classification. All evaluated techniques are described next.

\subsection{Deep Architectures} \label{sec:nns}

Convolutional Neural Networks (ConvNets)~\cite{goodfellow2016deep} have become the standard state-of-the-art technique for visual recognition over the last decade.
Their capability to provide end-to-end feature learning and inference turns them into powerful statistical models for computer vision applications, including scene classification.
Supported by this, we evaluated several ConvNet-based approaches for multi-view image classification using the proposed datasets.
All experimented techniques are described next.

\begin{figure*}[!ht]
    \centering
    \subfloat[AlexNet~\cite{alexnet} architecture] {
        \includegraphics[width=.95\columnwidth]{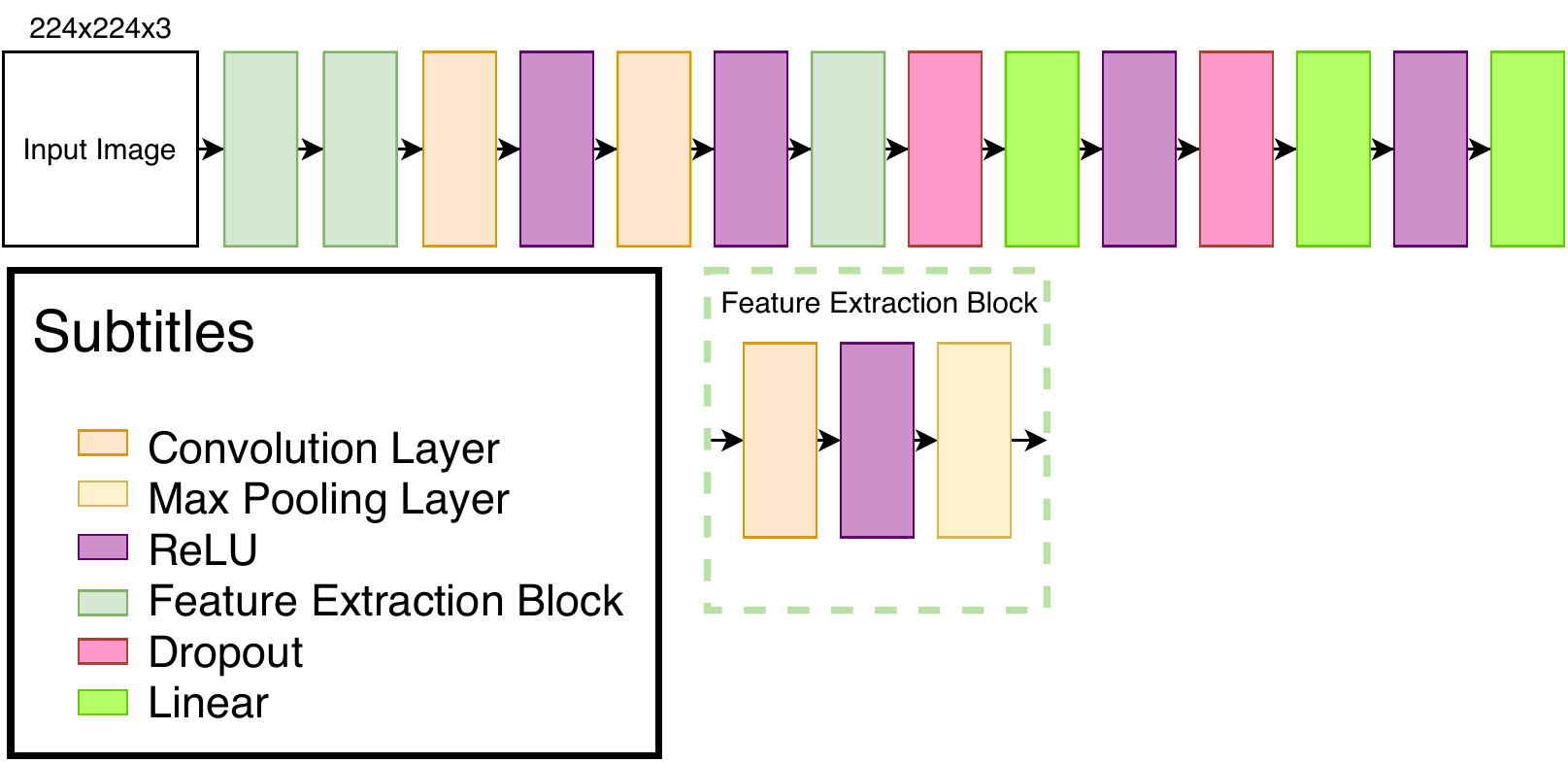}
        \label{fig:alexnetA}
    }
    \qquad
    \subfloat[VGG-11~\cite{vgg} architecture]{
        \includegraphics[width=.95\columnwidth]{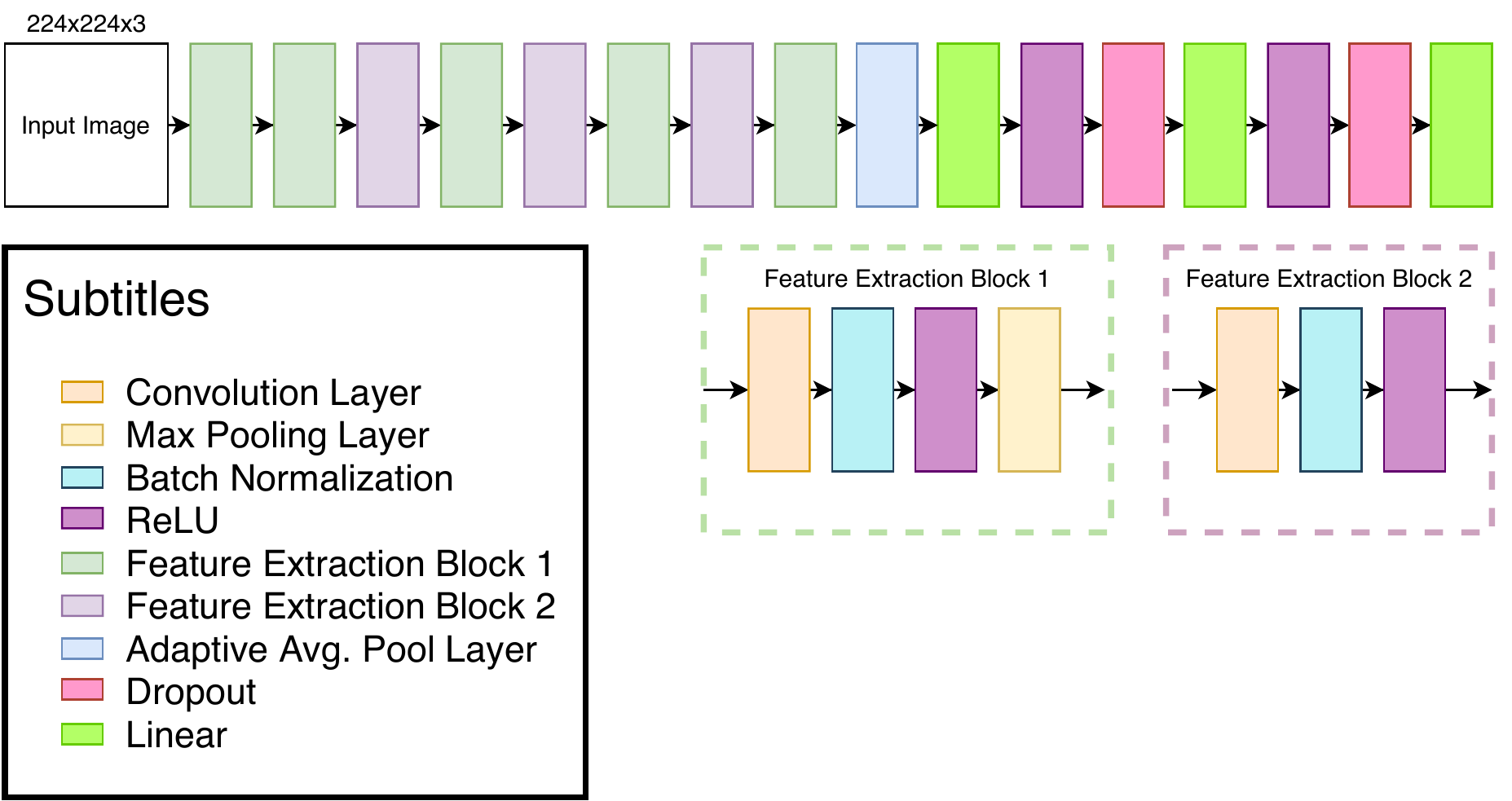} 
        \label{fig:VGGA}
    }
    \qquad
    \subfloat[Inception-V3~\cite{inceptionv3} architecture]{
        \includegraphics[width=.95\columnwidth]{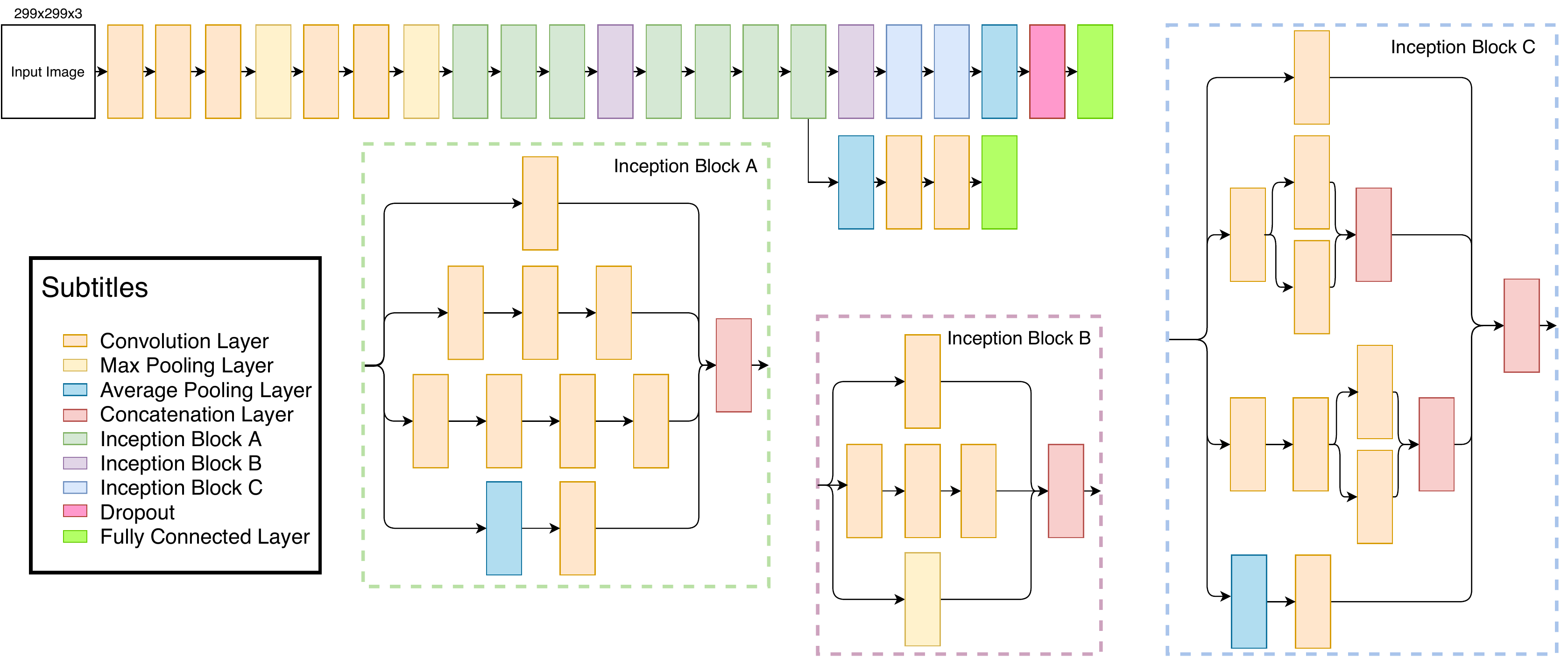}
        \label{fig:inceptionA}
    }
    \qquad
    \subfloat[ResNet-18~\cite{resnet} architecture]{
        \includegraphics[width=.95\columnwidth]{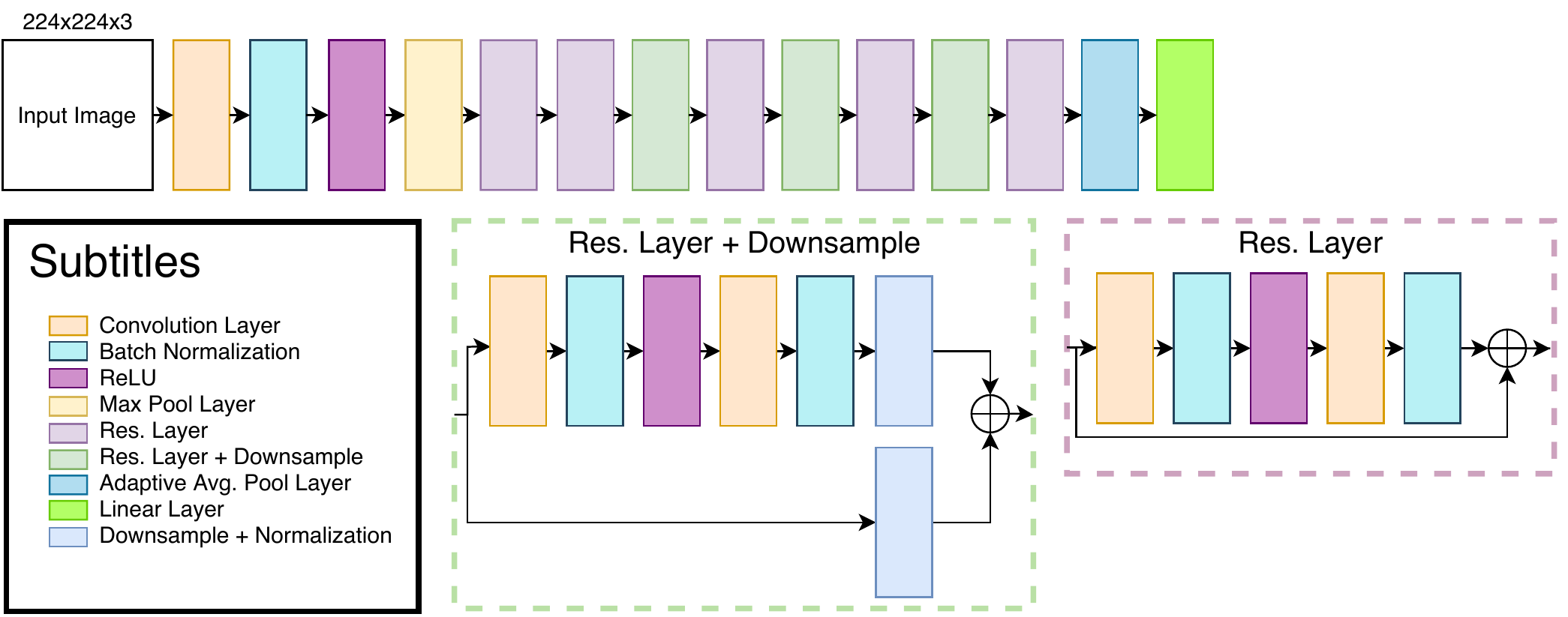}
        \label{fig:resnetA}
    }
    \qquad
    \subfloat[DenseNet-169~\cite{densenet} architecture]{
        \includegraphics[width=.95\columnwidth]{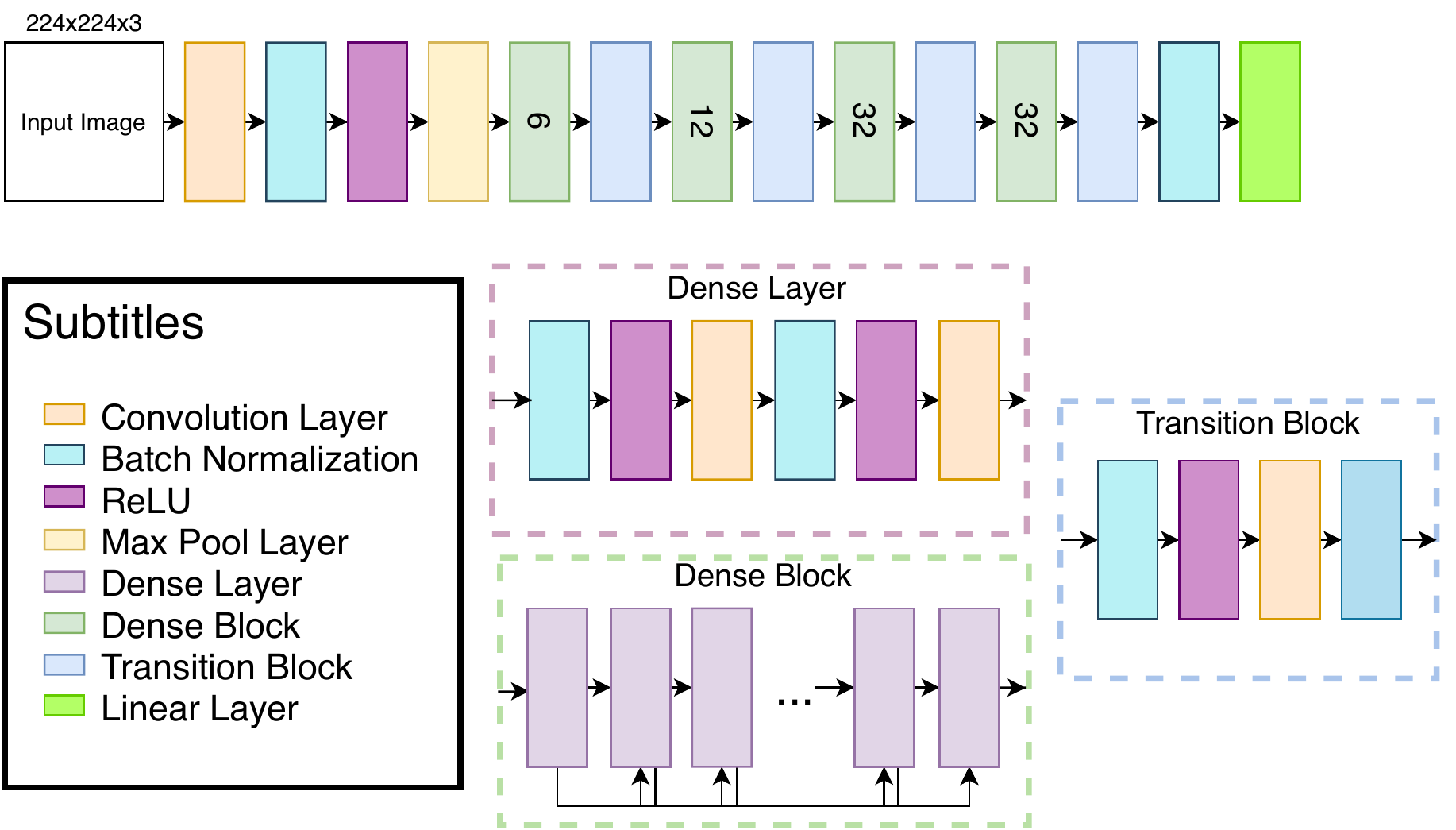}
        \label{fig:densenetA}
    }
    \qquad
    \subfloat[SqueezeNet~\cite{squeezenet} architecture]{
        \includegraphics[width=.95\columnwidth]{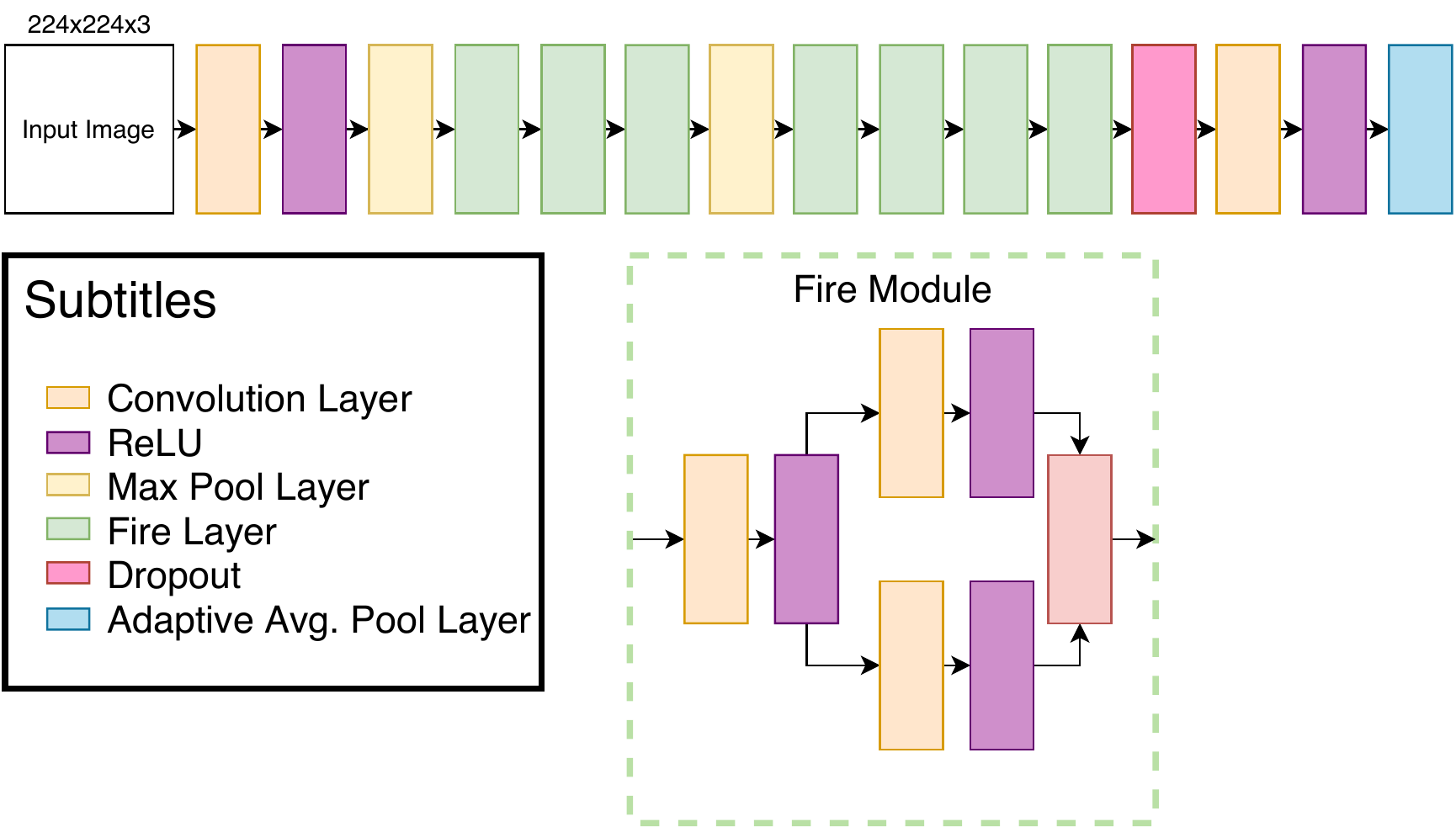} 
        \label{fig:squeezenetA}
    }
    \qquad
    \subfloat[Squeeze and Excitation ResNet-50~\cite{seresnet} architecture]{
        \includegraphics[width=.95\columnwidth]{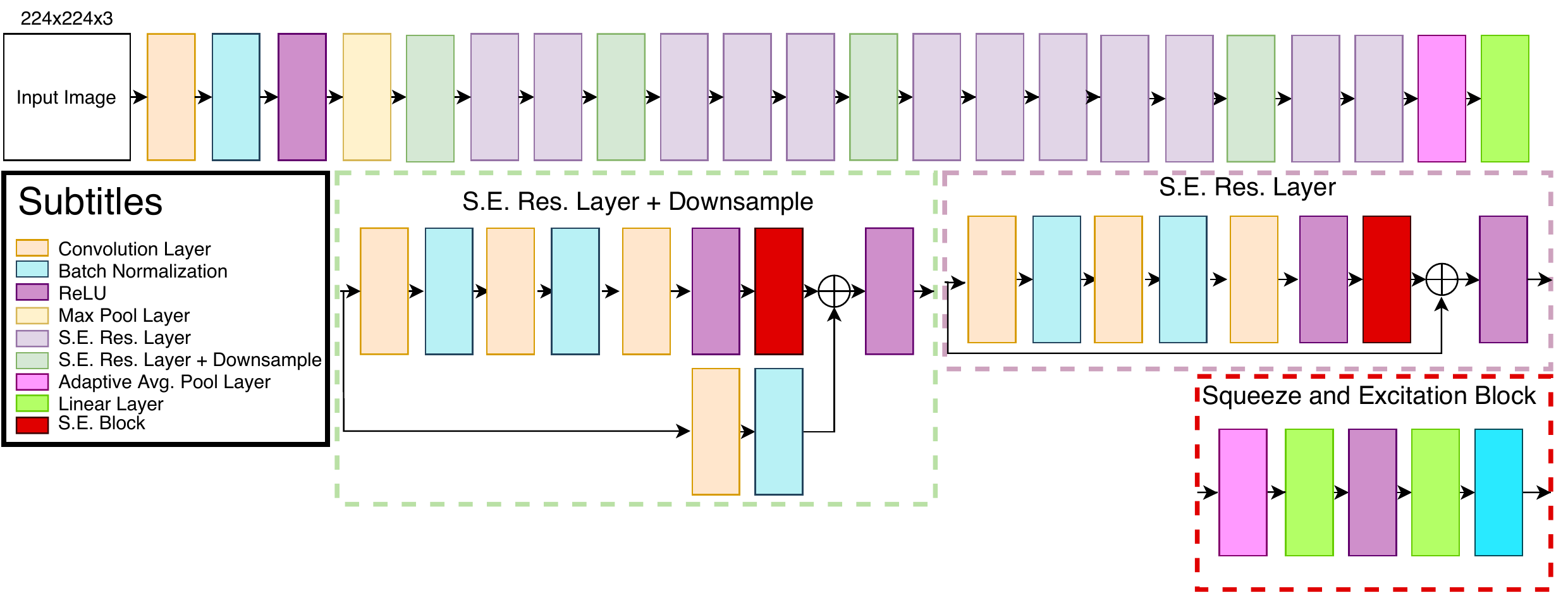} 
         \label{fig:seresnetA}    
    }
    \qquad
    \subfloat[Selective Kernels ResNet-101~\cite{sknet} architecture]{
        \includegraphics[width=.95\columnwidth]{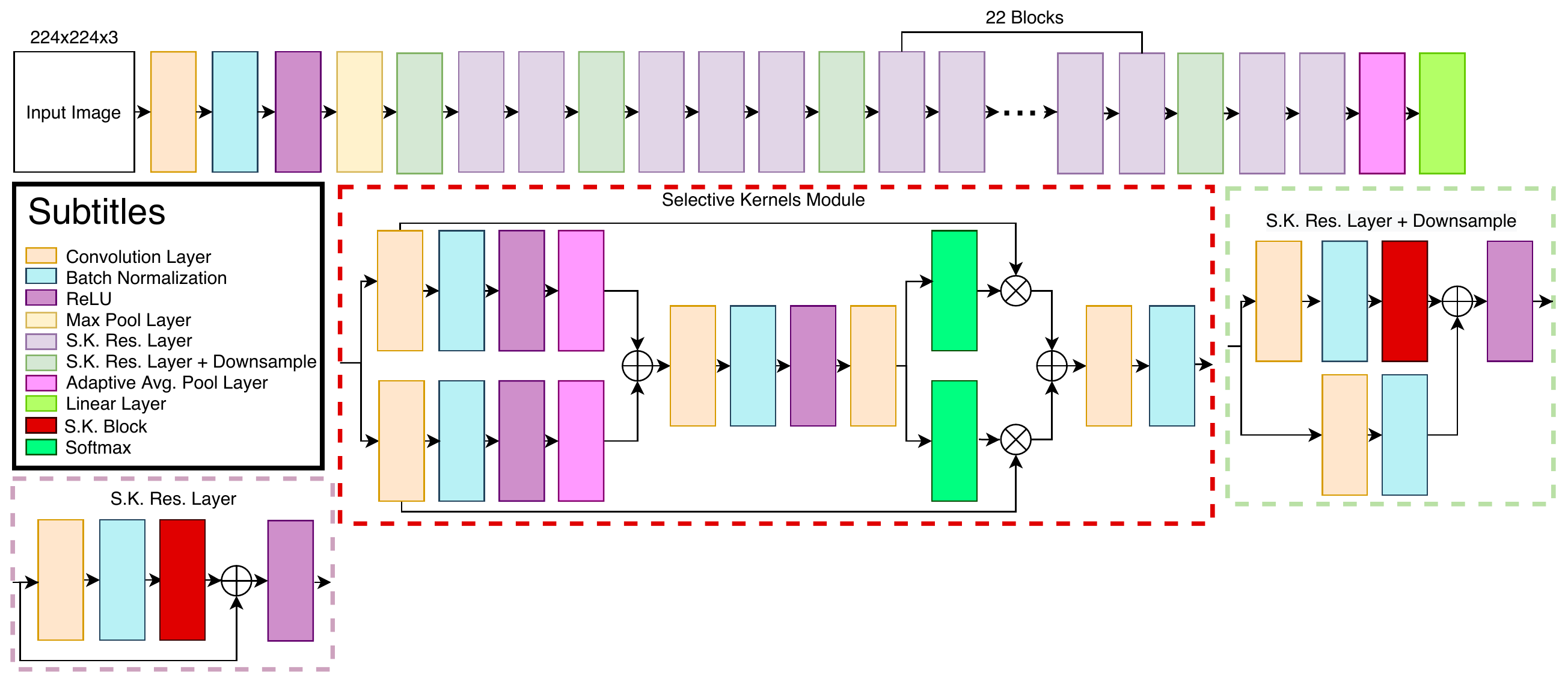} 
        \label{fig:sknetA}
    }
    \caption{Benchmarked architectures.}
   % \label{fig:inception}
\end{figure*}

\noindent \textbf{AlexNet}~\cite{alexnet}. 
The first network evaluated is the AlexNet one.
Originally proposed for and winner of the ILSVRC 2012 competition, this pioneer ConvNet is composed of five convolutional layers, some of which are followed by max-pooling layers, and three fully-connected layers with a final softmax.
The first convolutional layers use large convolutional filters in order to quickly reduce the spatial resolution of the input image. %Lastly, it is possible to see how the AlexNet architecture works looking at figure \ref{fig:alexnetA}.
Figure~\ref{fig:alexnetA} presents the architecture of AlexNet network. 

%\begin{figure*}[!ht]%
%    \centering
%    \subfloat[Original alexnet architecture.] {
%        \includegraphics[width=.95\columnwidth]{images/architectures/alexnet.pdf}
%        \label{fig:alexnetA}
%    }%
%    \qquad
%    \centering
%    \subfloat[Proposed early fusion architecture for  alexnet.]{
%        \includegraphics[width=.95\columnwidth]{images/architectures/alexnet_fusion.pdf}
%        \label{fig:alexnetB}
%    }%
%    \caption{Alexnet~\cite{alexnet} architectures.}%
%    \label{fig:alexnet}%
%\end{figure*}

%This network, proposed for classification of the ImageNet dataset~\cite{imagenet_cvpr09}, surpassed all other methods by a large margin in the ILSVRC 2012 competition.
%AlexNet~\cite{alexnet} used large convolutional filters in the first layers in order to quickly reduce the spatial resolution of the input image, followed by a sequence $3\times3$ convolutions and Fully Connected (FC) layers in the end of the network for inference.
%After some of the convolutional layers, 2D max pooling was employed, aiming to filter the regions with higher activations yielded from convolutions.

\noindent \textbf{VGG}~\cite{vgg}. 
This work was the first one to observe that smaller sequential convolutional filters had the representation capabilities of one single large trainable convolutional kernel.
Supported by this, the authors deepened the network, that has 8 $3\times3$ convolutional layers, 5 pooling ones and 4 fully-connected ones (considering the softmax). %Finally, looking at figure \ref{fig:VGGA}, it is possible to see how a VGG-11 architecture operates.
Figure~\ref{fig:VGGA} illustrates the architecture of VGG-11 network.

%\begin{figure*}[!ht]%
%    \centering
%    \subfloat[Original VGG-11 architecture.]{
%        \includegraphics[width=.95\columnwidth]{images/architectures/VGG.pdf} 
%        \label{fig:VGGA}
%    }%
%    \qquad
%    \centering
%    \subfloat[Proposed early fusion architecture for  VGG-11.]{
%        \includegraphics[width=.95\columnwidth]{images/architectures/VGG_fusion.pdf}
%        \label{fig:VGGB}
%    }%
%    \caption{VGG-11~\cite{vgg} architectures.}%
%    \label{fig:VGG}%
%\end{figure*}

\noindent \textbf{Inception}~\cite{inception_original,inceptionv3}.
Following the same guidelines of the VGG network~\cite{vgg}, this architecture employed more convolutional layers in order to increase the feature extraction ability.
Specifically, this network is based on the ``Inception'' modules that exploit feature diversity through parallel convolutions with different filter sizes.
This module is replicated several times producing the final architecture that has 48 layers. %Lastly, looking at figure \ref{fig:inceptionA}, it is possible to see how an inception-v3 architecture and inception modules work and how those blocks are integrated in the model.
Through the Figure \ref{fig:inceptionA}, it is possible to see how an inception-v3 architecture and inception modules work.

%%%%%%%%%%%%%%%%%%%%%%%%%%%%%%%%%%%%%%%%%%%%%%%%%%%%
%%%%%%%%%%%%%%%%%%%%%%%%%%%%%%%%%%%%%%%%%%%%%%%%%%%%

% \begin{figure*}[!ht]%
%     \centering
%     \subfloat[Original Inception v3 architecture.]{
%         \includegraphics[width=.95\columnwidth]{images/architectures/inception.pdf}
%         \label{fig:inceptionA}
%     }%
%     \qquad
%     \centering
%     \subfloat[Proposed early fusion architecture for inception v3.]{
%         \includegraphics[width=.95\columnwidth]{images/architectures/inception_fusion.pdf}
%          \label{fig:inceptionB}
%     }%
%     \caption{Inception v3~\cite{inceptionv3} architectures.}%
%     \label{fig:inception}%
% \end{figure*}

\noindent \textbf{ResNet}~\cite{resnet}. 
This work was the first one to notice that adding even more layers to the architecture only worsened the vanishing gradient problem.
%the upper limit for deepening ConvNets.
So, to mitigate this problem, the authors employed shortcut connections to allow the efficient training of earlier layers in the ConvNet.
Based on this concept, several networks were proposed, some of them with hundreds or even thousand convolutional layers.
In this work, ResNet-18~\cite{resnet}, which has 18 convolutional layers with adding shortcuts, was evaluated. 
Figure \ref{fig:resnetA} shows how a ResNet-18 architecture is built.

%\begin{figure*}[!ht]%
%    \centering
%    \subfloat[Original resnet-18 architecture.]{
%        \includegraphics[width=.95\columnwidth]{images/architectures/resnet.pdf}
%        \label{fig:resnetA}
%    }%
%    \qquad
%    \centering
%    \subfloat[Proposed early fusion architecture for  resnet-18.]{
%        \includegraphics[width=.95\columnwidth]{images/architectures/resnet_fusion.pdf}
%        \label{fig:resnetB}
%        }%
%    \caption{Resnet-18~\cite{resnet} architectures.}%
%    \label{fig:resnet}%
%\end{figure*}

%was introduced by~\cite{resnet}, and its main contribution is the reformulation of the layers of a deep neural network to learn residual functions that reference the input layers in order to handle the degradation problem of deep neural networks with many layers. The results consist of networks with deep layers that were easier to optimize and presented lower training error than their plain counterparts.

\noindent \textbf{DenseNet}~\cite{densenet}.
Following the same idea of the ResNets~\cite{vgg}, this architecture employed shortcut connections in order to allow the gradients to easily flow and better optimize the initial layers.
The difference between ResNets~\cite{vgg} and DenseNet~\cite{resnet} is that in the former one, the shortcuts add the inputs, while in the latter one, the input layers are concatenated in the shortcut connections.
Again, due to this shortcut design, dense architectures, with hundreds or even thousand convolutional layers, were proposed and employed in several applications~\cite{resnet}.
In this work, DenseNet-169~\cite{densenet}, which has 169 convolutional layers with shortcuts, was evaluated. The architecture of this model is presented in Figure \ref{fig:densenetA}.

%\begin{figure*}[!ht]%
%    \centering
%    \subfloat[Original densenet-169 architecture.]{
%        \includegraphics[width=.95\columnwidth]{images/architectures/densenet.pdf}
%        \label{fig:densenetA}
%    }%
%    \qquad
%    \centering
%    \subfloat[Proposed early fusion architecture for densenet-169.]{
%        \includegraphics[width=.95\columnwidth]{images/architectures/densenet_fusion.pdf}
%        \label{fig:densenetB}
%    }%
%    \caption{Densenet-169~\cite{densenet} architectures.}%
%    \label{fig:densenet}%
%\end{figure*}

\noindent \textbf{SqueezeNets}~\cite{squeezenet}.
This network uses a combination of pruning, compression techniques, and fire modules composed of squeeze and expand convolutions in order to create a lean and efficient architecture that can be incorporated into devices with limited memory (such as mobile).
In fact, SqueezeNets are able to achieve visual recognition objective scores close to early ConvNet architectures (as AlexNet~\cite{alexnet}) with between one or two orders of magnitude fewer parameters.
Figure \ref{fig:squeezenetA} illustrates how a fire module works and how they are integrated with a SqueezeNet architecture.

%\begin{figure*}[!ht]%
%    \centering
%    \subfloat[Original squeezenet architecture.]{
%        \includegraphics[width=.95\columnwidth]{images/architectures/squeezenet.pdf} 
%        \label{fig:squeezenetA}
%    }%
%    \qquad
%    \centering
%    \subfloat[Proposed early fusion architecture for  squeezenet.]{
%        \includegraphics[width=.95\columnwidth]{images/architectures/squeezenet_fusion.pdf}
%        \label{fig:squeezenetB}
%    }%
%    \caption{Squeezenet~\cite{squeezenet} architectures.}%
%    \label{fig:squeezenet}%
%\end{figure*}

\noindent \textbf{Squeeze and Excitation Networks}~\cite{seresnet}.
Instead of focusing on spatial components to enhance feature extraction results, this work focus on the relationship between channels. 
For this task, the authors propose a new block named ``Squeeze and Excitation block''. 
This block operates recalibrating channel-wise feature impacts by modeling interdependencies between those channels in an explicit way.
In this work, the authors also show that using these blocks, networks can outperform previously state-of-the-art results on the ImageNet dataset \cite{imagenet_cvpr09} and that the use of this block can be easily adapted to existing architectures.
Figure \ref{fig:seresnetA} shows how a ``Squeeze and Excitation block'' works and how it can be implemented in a ResNet-50 architecture.

% \begin{figure*}[!ht]%
%     \centering
%     \subfloat[Original squeeze and excitation resnet-50 architecture.]{
%         \includegraphics[width=.95\columnwidth]{images/architectures/seresnet.pdf} 
%          \label{fig:seresnetA}    
%     }%
%     \qquad
%     \centering
%     \subfloat[Proposed early fusion architecture for squeeze and excitation resnet-50.]{
%         \includegraphics[width=.95\columnwidth]{images/architectures/seresnet_fusion.pdf}
%          \label{fig:seresnetB}
%     }%
%     \caption{Squeeze and excitation resnet-50~\cite{seresnet} architectures.}%
%     \label{fig:seresnet}%
% \end{figure*}

\noindent \textbf{Selective Kernels Networks}~\cite{sknet}. Most of the designed ConvNets use receptive fields of the same size in each one of its layers. 
In this work, the authors propose an attention block named ``Selective Kernel unit''. 
The main objective of this block is to allow each neuron to adaptively adjust the size of its receptive field, looking at different scales of input information.
Relating to the functioning of this block, it is based on a fusion of kernels that have different sizes using a softmax attention, guided by the input information that enters the block.
In this work, the authors also show that using these blocks on a ResNet \cite{resnet} can outperform previously state-of-the-art results on the ImageNet dataset \cite{imagenet_cvpr09}.
Through Figure \ref{fig:sknetA} it is possible to see how a ``Selective Kernel unit'' operates and how it was integrated into a ResNet-101 architecture.

\subsection{Fusion Methods} \label{sec:fusion-methods}

To enhance scene classification results, we evaluated several late fusion algorithms and a novel early fusion methodology. 
In this work, those techniques were applied to fuse aerial/ground/satellite features, acquiring new information, and enhancing the final predictions. In the sections bellow, we will describe all such techniques.

\subsubsection{Early Fusion Methods} \label{sec:early-fusion-methods}

In order to exploit the correlations and interactions between low level features from different modalities \cite{baltruvsaitis2018multimodal}, we propose early fusion approaches based on the deep architectures used for the experiments. 
A great advantage of early fusion approaches is that they require the training of a single model, which usually results in compacted models compared to the late fusion ones.

\begin{figure}[!ht]
	\centering
	\includegraphics[width=0.95\columnwidth]{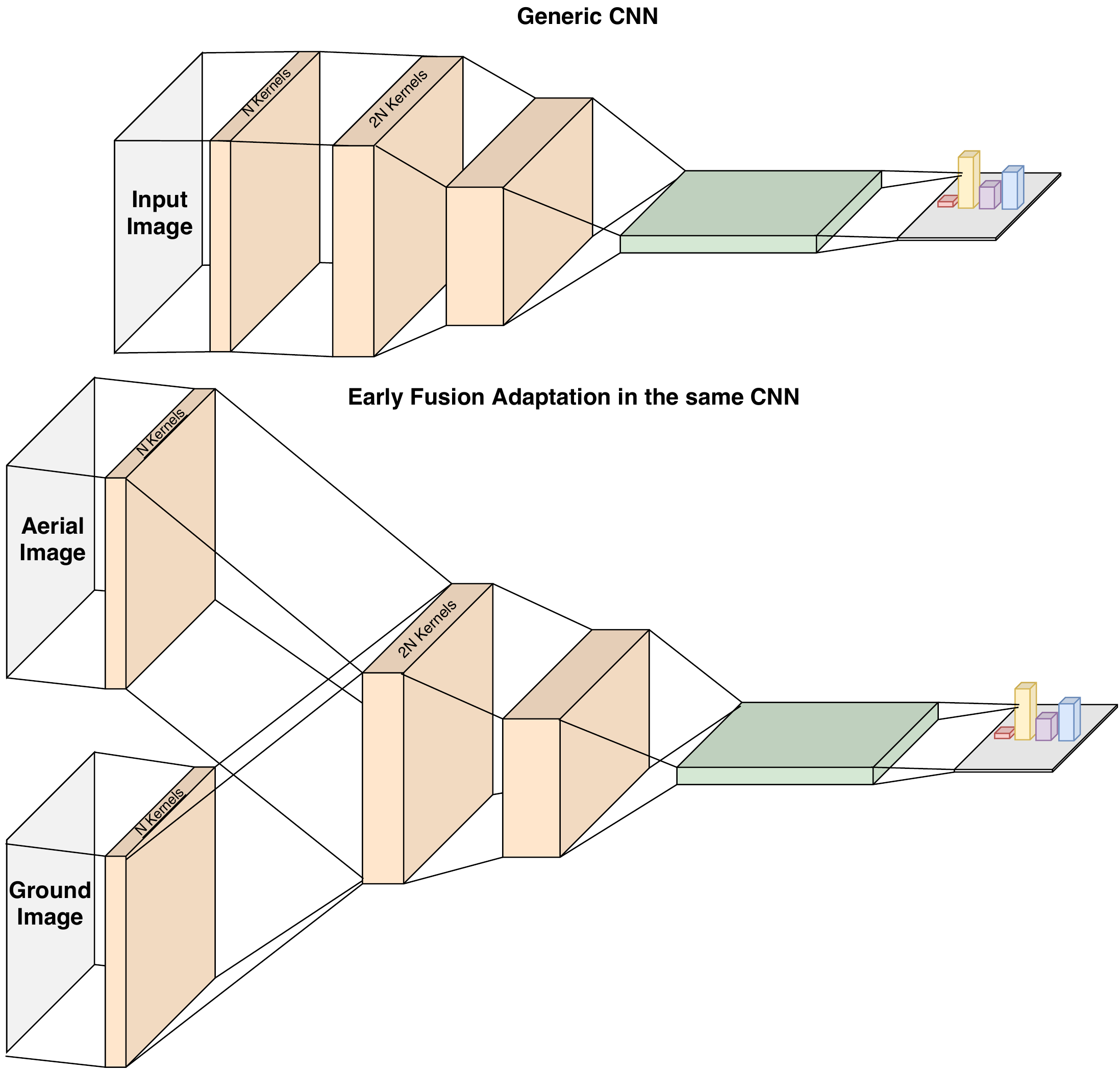}
	\caption{Example of the proposed early fusion methodology.}
	\label{fig:earlyincep}
\end{figure}

The early fusion strategy proposed for this work consists of using the first feature extraction layers of the target network as a backbone. This backbone is replicated to aerial and ground images. 
The fusion of the features is made by applying a concatenation layer on the low-level features, which results in a tensor that contains the double amounts of kernels than the original ones. 
The choice of where those concatenations were performed is based on the total of kernels that each convolution layer have. 
So, to be possible of fully explore pre-trained models, we decided to concatenate those feature vectors before the first convolution layer that doubles its amount of kernels in the target network.
In this way, we ignore the convolution that duplicates this amount of kernels and substitute it to a fusion that also duplicates the amount of feature vectors, as can be seen in Figure \ref{fig:earlyincep}.

%As said before, some of the great advantages of early fusion models comparing to late fusion ones are related to the sizes of the models and the necessary time to train them. Looking at Table \ref{tab:times}, we can clearly see those advantages.

% \begin{figure*}[!ht]
% 	\centering
% % 	\includegraphics[width=.96\textwidth]{images/vgg-imp.pdf}
% 	\includegraphics[width=.95\textwidth]{images/architectures/earlyfusion_process2.pdf}
% 	\caption{Example of the proposed early fusion strategy.}
% 	\label{fig:earlyincep}
% \end{figure*}

\subsubsection{Late Fusion Methods} \label{sec:late-fusion-methods}

\begin{figure}[!ht]
	\centering
	   
	   \includegraphics[width=0.95\columnwidth]{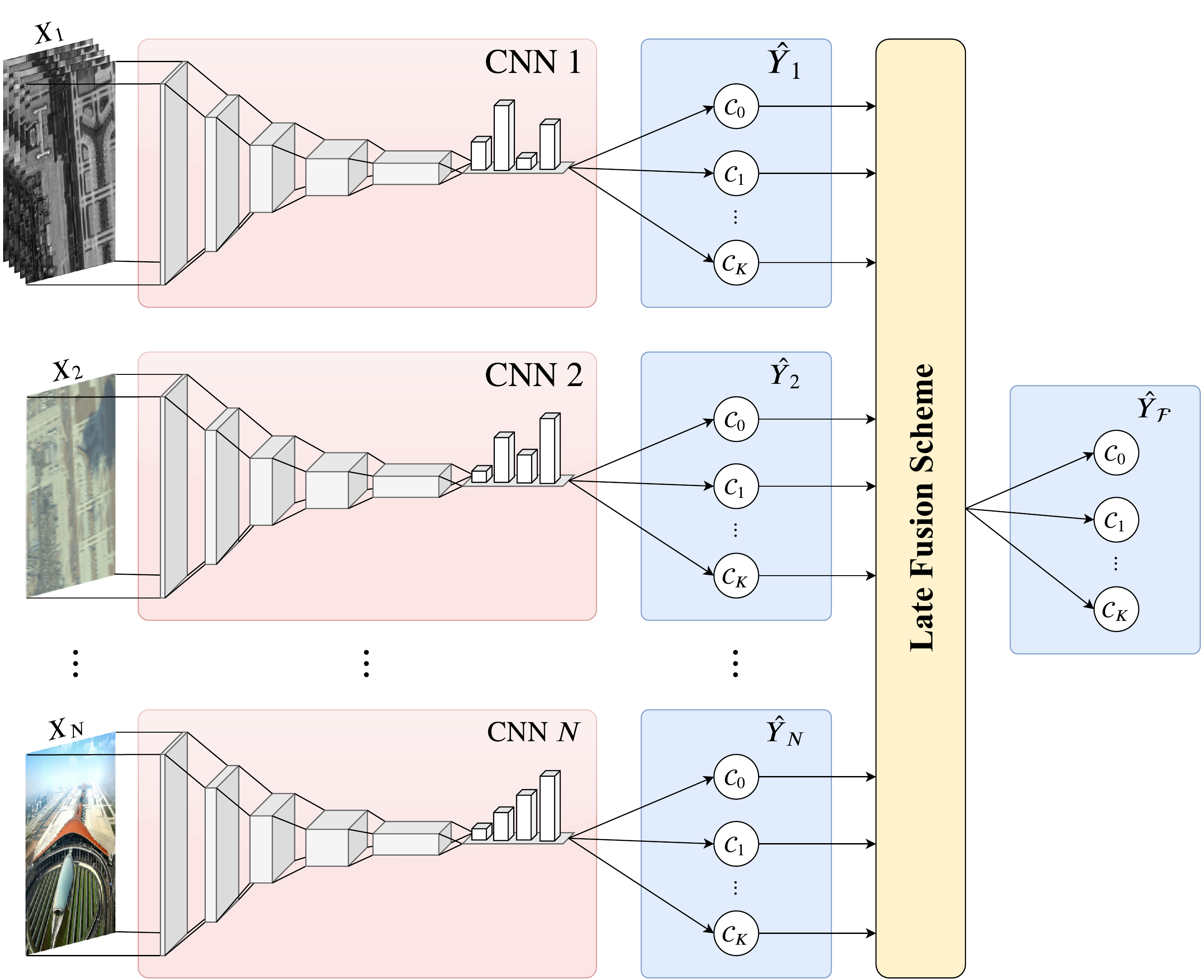}
	\caption{A typical late fusion pipeline. As can be seen, each ConvNet is trained individually and their outputs are combined using a late fusion algorithm, resulting in the final prediction.}
	\label{fig:pipeline}
\end{figure}

Late fusion or decision-based algorithms performs integration of results after each of the modalities has made a prediction~\cite{baltruvsaitis2018multimodal}.
Those algorithms use uni-modal decision values and combine them using different types of fusion mechanisms, such as averaging, voting schemes, or weighting based. Figure \ref{fig:pipeline} presents a typical late fusion procedure exploited in this work.

To formally define all the fusion operations used for this work, for all definitions of this section we will use the following notation.
Let $\sigma_{i}$ be the softmax scores returned by the network $i$, $\alpha_{i}$ be the accuracy score that the network $i$ achieved on the validation set of each dataset and $m$ be the number of networks used to perform a fusion operation.

\noindent \textbf{Sum}. The sum fusion is a well known late fusion algorithm. The main idea of it is to sum all the vectors (softmax scores) and select the index which contains the maximum value of this sum as the prediction. This procedure is formally defined by equation \ref{eq1:sum}.

\begin{equation}
    Sum_{Prediction}~=~\argmax~ \sum_{i=1}^{m} \sigma_{i}
    \label{eq1:sum}
\end{equation}

\noindent \textbf{Majority Voting}. The majority voting fusion is another well-known late fusion method in the literature.
This method is based on the concept of a democratic election, i.e., each model act as a voter and provides its output as a vote, the final prediction is selected as the returned value with more votes.
To mathematically express this procedure, it was used a mode operation, that is a statistic that indicates the most common element contained in a vector. The majority voting procedure is properly defined by equation \ref{eq1:mv}.

\begin{equation}
    MV_{Prediction}~=~\mode~\argmax~\sigma_{i},~\forall~i~\in~[1,m]  
    \label{eq1:mv}
\end{equation}
In the case of this work, majority voting was used to perform late fusion between 2 or 3 models, so ties constantly happen. 
To solve this issue, as a tiebreaker it was used the confidence (probability value) that each model has on its answer. 
In this way, when a vote ties, we select the output with the biggest confidence between all models' outputs. The same process was used for fusions using only 2 models since there is no point in checking what is the most common vote between two voters. The procedure used for tiebreaker and voting using only 2 models is formally defined by equation \ref{eq2:mv}.

\begin{equation}
\begin{aligned}
& MV_{Prediction} = \beta_{\theta}, where \\   
    & \beta_{i}~=~\argmax~\sigma_{i} \\ 
    & \theta~=~\argmax~\max~\sigma_{i},~\forall~i~\in~[1,m]
    \label{eq2:mv}   
\end{aligned}
\end{equation}

\noindent \textbf{Weighted Sum}. As can be noted by its name, this method operates in a similar way that sum fusion does. 
The main difference between them is that the weighted sum multiplies values (weights) while it is performing a sum operation. This procedure is useful in situations that different classifiers have very distinct results. 
So, in this case, the weighted sum can use values for trying to calibrate this huge variance between the models' results.
The formal definition of weighted sum can be checked in equation \ref{eq:wsum}, in which the weights used for the experiments on this work were taken from the individual performance score (accuracy) of each model on the validation set of each dataset. 

\begin{equation}
    WSum_{Prediction}~=~\argmax~ \sum_{i=1}^{i=m} \alpha_{i} \sigma_{i}
    \label{eq:wsum}
\end{equation}

\noindent \textbf{Minimum}. The main advantage of the minimum fusion method is that the algorithm can eliminate possible overfitting that may have occurred during the training phase. 
The first step of the method is to select the individual prediction of each model (the index which contains the maximum value on softmax scores vector).
After that, the method looks for the scores associated to each one of the returned indexes and returns as the final prediction the index that has the smallest score associated with it.
Equation \ref{eq:min} formally defines the described procedure.

\begin{equation}
\begin{aligned}
    & Min_{Prediction}~=~\beta_{\theta},~where \\   
    & \beta_{i}~=~\argmax~\sigma_{i}, \\ 
    & \theta~=~\argmin~\max~\sigma_{i},~\forall~i~\in~[1,m]
    \label{eq:min}   
\end{aligned}
\end{equation}

\noindent \textbf{Product}. The product fusion is a very used late fusion algorithm. The main idea of it is to perform an element-wise multiplication between softmax scores and after that return the index which contains the biggest value. This procedure is defined by equation \ref{eq:prod}. 
\begin{equation}
    Prod_{Prediction}~=~\argmax~ \prod_{i=1}^{m} \sigma_{i}
    \label{eq:prod}
\end{equation}

\section{Experimental Setup}
\label{sec:experimental_setup}

% \todo{REVISORES:
% -What about data augmentation in terms of flipping, rotating, increasing the level of noise, etc. to obtain increased robustness ?
% }

In this section, we describe the experimental setup used for the experiments using both datasets.
It is important to mention that all the methods, previously described in Section~\ref{sec:methods}, were used for the experiments, and in all of those experiments a 5-fold cross-validation protocol was used to properly evaluate each technique. We reported the mean of balanced accuracy and/or F1-score, taken from all $5$ folds experiments with its correspondent standard deviation. 
Finally, in Section~\ref{sec:airound} we present the protocol used to train the models using \thedataset~dataset, while in Section~\ref{sec:cvbrct} we detail the methodology used for the CV-BrCT dataset models.

%to perform the aerial/ground scene classification. %while Section~\ref{sec:uda-setup} introduces the protocol employed to perform aerial/ground UDA.
%It is important to mention that all the experiments uses \thedataset ~as dataset.

%Finally, relating to the hardware setup, all the experiments were performed on a 64-bit Intel i9 7920X machine with 64GB of RAM memory and a GeForce RTX 2080 TI.

\subsection{\thedataset}
\label{sec:airound}

Since all the networks used for this work are well known in the literature, it is possible to find pre-trained models of them. So, in order to allow a better comparison and understanding of the most suitable training strategy for \thedataset, we trained all the models from scratch and fine-tuned.
For all the experiments made on \thedataset, each model was trained for $300$ epochs, using early stop technique with $30$ epochs checking for improvements in validation. Relating to the other hyperparameters, it was used a batch size of $32$, stochastic gradient descent as optimizer, a learning rate of $0.001$, and a momentum of $0.9$. 
Finally, about the data augmentation techniques applied, it was performed the randomized crop and random horizon flip. 

Relating to the models evaluated, for late fusion, we trained a individual network for each kind of data, as was shown in Figure~\ref{fig:pipeline}. For a better comparison, we evaluated the combinations of the models using all the late fusion algorithms previously described. 
In order to test all possible combinations of fusions between different views, each late fusion algorithm fuses outputs of $2$ or $3$ networks, trained in different views, by combining alternated models' outputs, e.g., 3-views, aerial with ground, etc. 
All the combinations were made using models with the same architecture and trained using data from only 1-view perspective.

For the early fusion models, an end-to-end training was performed, using aerial + ground paired data as inputs. 
All the training process was also made using the same combination of hyperparameters used to train the 1-view individual models.

% In this scheme, ConvNets trained individually have their outputs combined using a late fusion algorithm, resulting in the final prediction.
% Several fusion approaches~\cite{alpaydin2014introduction} were considered, including: 
% (i) sum; 
% (ii) majority voting; 
% (iii) weighted sum; 
% (iv) minimum; and 
% (v) product. 

%It is important to highlight that, in some of these methods, minor adaptations were required.
%Specifically, the weighted sum fusion used the kappa score of the baseline experiments to calculate the weight value used to perform the sum for each classifier.
%In this way, the trust given for each classifier is based on how well this classifier performed its predictions for each view.
%In the majority voting scheme, the confidence given by each model was employed as a tiebreaker. 
%In this way, when a voting ties, we select the prediction from only one of the models. This selection is made based on how confident the classifier is on its answer, i.e., we select this prediction based on the greater value presented in the softmax scores between all voters.

\subsection{CV-BrCT}
\label{sec:cvbrct}

For the second dataset, we used a very similar protocol than the one previously described. The main differences between them are that we trained each model for $100$ epochs, instead of $300$, and we used an early stop with $10$ epochs. This decision was made because the CV-BrCT has way more samples than \thedataset~and the models tended to converge faster. Naturally, for this dataset we could not perform the same set of late fusions, because it does not have Sentinel-2 data, so we performed the late fusions only using aerial and ground data.

\section{Results and Discussion}
\label{sec:results}

In this section, we present and discuss the obtained results. 
The Results for \thedataset~dataset are presented in Section \ref{sec:result_air}. 
Sections \ref{sec:networks_air} and \ref{sec:fusion_air} present the results achieved for training networks using only 1-view type and applying fusion techniques, respectively.
Relating to CV-BrCT dataset, the results can be found in Section \ref{sec:result_cv}.
Following the same organization used for \thedataset, we present the results of the models trained using 1-view in Section \ref{sec:networks_cv}, while the results of the models using fusion techniques can be found in Section \ref{sec:fusion_cv}.

\subsection{Experiments on the \thedataset} \label{sec:result_air}

%%%%%%%%%%%%%%%%%%%%%%%%%%%%%%%%%%%%%%%%%%%%%%%%%%%%%%%%%%%%%%%%
\subsubsection{Networks Architectures Comparison} \label{sec:networks_air}

%Results for aerial/ground scene classification were organized to address three research questions:
%(i) which network/training strategy is the most suitable for aerial/ground image classification in the context of the proposed dataset?
%(ii) can the fusion of aerial/ground information improve the general performance of the networks scene classification?
%(iii) which classes and points of view benefit more from feature fusion?

%%%%%%%%%%%%%%%%%%%%%%%%%%%%%%%%%%%%%%%%%%%%%%%%%%%%%%%%%%%%%%%%
%\subsubsection{ConvNet Architectures} \label{subsec:baseline}

Here, we present the results obtained from the deep learning-based models trained individually, for each view, from scratch and fine-tuned (from the ImageNet dataset~\cite{imagenet_cvpr09}).
As introduced, the objective is to analyze and define the most suitable network and training strategy for \thedataset~dataset.
All obtained results are presented in Table~\ref{tab:baseline-airound}.
It is important to highlight we did not fine-tune the networks for sentinel-2 images, given the incompatibility between the number of bands of this data and the number of input channels expected by the networks, i.e., given that Sentinel-2 data has 13 channels, and the first convolution layer of the evaluated architectures receive only 3 input bands (RGB).
We also decided not to use only Sentinel-2's RGB channels as input to pre-trained models, because the average spatial resolution of the aerial-RGB images is much better than Sentinel-2, providing much more details and information.

Analyzing the results, it is possible to observe that, as 
has been seen in the literature, fine-tuned networks produced better outcomes than their counterpart models trained from scratch~\cite{nogueira2017towards}.
Comparing each training strategy separately, we can notice that, in both cases, the networks yielded very similar results (mainly for the aerial and ground data) without one model clearly outperforming others.
%We believe that this is motivated by the complexity of the dataset, which does not allow a method to stand out without using multiple sources of information.
Furthermore, considering the distinct input data, one may note, from the experiments with networks trained from scratch, that aerial and ground images tend to produce comparable outcomes, while sentinel-2 data tend to yield worse results.
This may be justified by the difference in the spatial resolution of the images since Sentinel-2 data has a resolution in meters per pixel whereas the aerial and ground have pixel resolution in centimeters per pixel.

\begin{table}[!ht]
\centering
\setlength{\tabcolsep}{5pt}
\begin{adjustbox}{max width=\columnwidth}
\begin{tabular}{@{}ccccc@{}}
\toprule
\multirow{3}{*}{\textbf{\begin{tabular}[c]{@{}c@{}}Training\\ Strategy\end{tabular}}} & \multirow{3}{*}{\textbf{Network}} & \multicolumn{3}{c}{\textbf{Input Data}} \\ \cmidrule(l){3-5} 
 &  & \multicolumn{1}{c}{\textbf{Aerial}} & \multicolumn{1}{c}{\textbf{Ground}} & \multicolumn{1}{c}{\textbf{Sentinel-2}} \\ %\cmidrule(l){3-5}
 \midrule
\multirow{8}{*}{\textbf{\begin{tabular}[c]{@{}c@{}}Training\\ from\\ scratch\end{tabular}}}
 & \textbf{AlexNet~\cite{alexnet}} & $0.75 \pm 0.06$ & $0.75 \pm 0.03$ & $0.51 \pm 0.07$ \\
 & \textbf{VGG~\cite{vgg}} & $\mathbf{0.79 \pm 0.07}$ & $\mathbf{0.78 \pm 0.03}$ & $\mathbf{0.71 \pm 0.06}$ \\
 & \textbf{Inception~\cite{inceptionv3}} & $0.69 \pm 0.07$ & $0.70 \pm 0.05$ & $0.69 \pm 0.07$ \\
 & \textbf{ResNet~\cite{resnet}} & $0.76 \pm 0.05$ & $0.75 \pm 0.04$ & $0.70 \pm 0.05$ \\
 & \textbf{DenseNet~\cite{densenet}} & $0.73 \pm 0.04$ & $0.74 \pm 0.03$ & $\mathbf{0.71 \pm 0.05}$ \\
 & \textbf{SqueezeNet~\cite{squeezenet}} & $0.70 \pm 0.07$ & $0.73 \pm 0.03$ & $0.59 \pm 0.04$ \\
 & \textbf{SENet~\cite{seresnet}} & $0.69 \pm 0.04$ & $0.69 \pm 0.07$ & $0.68 \pm 0.04$ \\
 & \textbf{SKNet~\cite{sknet}} & $0.67 \pm 0.05$ & $0.65 \pm 0.10$ & $0.60 \pm 0.04$ \\ \midrule
\multirow{8}{*}{\textbf{\begin{tabular}[c]{@{}c@{}}Fine tuning\\ from\\ ImageNet\end{tabular}}}
 & \textbf{AlexNet~\cite{alexnet}} & $0.82 \pm 0.05$ & $0.82 \pm 0.02$ & - \\
 & \textbf{VGG~\cite{vgg}} & $0.88 \pm 0.03$ & $0.86 \pm 0.03$ & - \\
 & \textbf{Inception~\cite{inceptionv3}} & $0.88 \pm 0.04$ & $\mathbf{0.88 \pm 0.03}$ & - \\
 & \textbf{ResNet~\cite{resnet}} & $0.87 \pm 0.02$ & $0.86 \pm 0.05$ & - \\
 & \textbf{DenseNet~\cite{densenet}} & $0.88 \pm 0.05$ & $0.87 \pm 0.02$ & - \\
 & \textbf{SqueezeNet~\cite{squeezenet}} & $0.85 \pm 0.05$ & $0.83 \pm 0.02$ & - \\
 & \textbf{SENet~\cite{seresnet}} & $0.87 \pm 0.02$ & $0.87 \pm 0.03$ & - \\
 & \textbf{SKNet~\cite{sknet}} & $\mathbf{0.90 \pm 0.02} $ & $0.87 \pm 0.01$ & - \\ \bottomrule
\end{tabular}
\end{adjustbox}
\caption{Results in terms of F1 Score of the evaluated models for \thedataset~dataset.}
\label{tab:baseline-airound}
\end{table}

%Concerning the first research question, we can conclude that fine-tuning is the best training strategy for using only one view of the proposed dataset.
%In terms of the best model, all networks produced very similar results (mainly for aerial and ground data) and could be further exploited.

%%%%%%%%%%%%%%%%%%%%%%%%%%%%%%%%%%%%%%%%%%%%%%%%%%%%%%%%%%%%%%%%
\subsubsection{Multi-View Fusion Strategies} \label{sec:fusion_air}

This section presents and discusses the results obtained applying early and late fusion techniques. 
%Since we stated that one of the great advantages of early fusion models comparing to late fusion ones are related to the sizes of the models and the necessary time to train them. 
%We evaluated a series of experiments, measuring the spent time to train each one of the models in both datasets. Table~\ref{tab:times} contain those results, demonstrating clearly those aforementioned advantages for \thedataset~dataset. 

\noindent \textbf{\thedataset~Early Fusion}. Like was specified before, for early fusion experiments we followed the scheme described at Section~\ref{sec:early-fusion-methods} using all the $8$ architectures that were also previously described.
In Table~\ref{tab:results-airound-early}, it is notable that all the results achieved a superior mark compared to the 1-view results reported in Table~\ref{tab:baseline-airound}.

%\todo{POLIR PARAGRAFO}
Analyzing the main fusion gains, for the models trained from scratch, we can highlight two models: the VGG~\cite{vgg}, that achieve the best results, and the inception-v3~\cite{inceptionv3}, which achieved the biggest gain in F1 score ($0.1$) comparing to the 1-view experiments, previously reported. 
For the fine-tuned models, we can highlight the DenseNet~\cite{densenet} adaption for early fusion, that obtained a gain of $0.05$ in F1 Score, comparing to the same network trained on single-view. 
It is also notable that early fusion DenseNet achieved the best overall result between all early fusion networks. 

Finally, for some results, it is notable that a downgrade occurred, if compared to the 1-view baseline presented in Table~\ref{tab:baseline-airound}. 
For those results, we hypothesize that the same feature degradation phenomenon\footnote{A destructive effect that occurs in training phase due to the misalignment of the geometry of the bottleneck features of the two image types.}, that was reported in~\cite{hoffmann2019model}, occurred.

\begin{table}[!ht]
\centering
\begin{adjustbox}{max width=\columnwidth}
\begin{tabular}{@{}ccccc@{}}
\toprule
\multirow{4}{*}{\textbf{\begin{tabular}[c]{@{}c@{}}Early Fusion\\ Networks\end{tabular}}} & \multicolumn{4}{c}{\textbf{Training Strategy}} \\ \cmidrule(l){2-5} 
 & \multicolumn{2}{c}{\textbf{From Scratch}} & \multicolumn{2}{c}{\textbf{Fine Tuning}} \\ \cmidrule(l){2-5} 
 & \textbf{B. Acc.} & \textbf{F1 Score} & \textbf{B. Acc.} & \textbf{F1 Score} \\ \midrule
\textbf{AlexNet~\cite{alexnet}} & $0.73 \pm 0.06$ & $0.80 \pm 0.06$ & $0.85 \pm 0.05$ & $0.89 \pm 0.03$ \\
\textbf{VGG~\cite{vgg}} & $\mathbf{0.76 \pm 0.03}$ & $\mathbf{0.83 \pm 0.03}$ & $0.83 \pm 0.04$ & $0.88 \pm 0.03$ \\
\textbf{Inception~\cite{inceptionv3}} & $0.71 \pm 0.06$ & $0.79 \pm 0.03$ & $0.87 \pm 0.02$ & $0.91 \pm 0.01$ \\
\textbf{ResNet~\cite{resnet}} & $0.69 \pm 0.04$ & $0.77 \pm 0.02$ & $0.85 \pm 0.03$ & $0.89 \pm 0.03$ \\
\textbf{DenseNet~\cite{densenet}} & $0.68 \pm 0.08$ & $0.77 \pm 0.05$ & $\mathbf{0.89 \pm 0.04 }$ & $\mathbf{0.94 \pm 0.01}$ \\
\textbf{SqueezeNet~\cite{squeezenet}} & $0.64 \pm 0.07$ & $0.75 \pm 0.06$ & $0.78 \pm 0.05$ & $0.85 \pm 0.03$ \\
\textbf{SENet~\cite{seresnet}} & $0.69 \pm 0.09$ & $0.77 \pm 0.06$ & $0.85 \pm 0.04$ & $0.89 \pm 0.02$ \\
\textbf{SKNet~\cite{sknet}} & $0.66 \pm 0.07$ & $0.74 \pm 0.04$ & $0.87 \pm 0.06$ & $0.91 \pm 0.04$ \\ \bottomrule
\end{tabular}
\end{adjustbox}
\caption{Results of the evaluated early fusion networks for \thedataset~dataset.}
\label{tab:results-airound-early}
\end{table}

\noindent \textbf{\thedataset~Late Fusion}. %After analyzing the performance of the deep learning-based approaches over each specific point of view from the proposed dataset, we conducted experiments to investigate if the combination of the multiple source data could improve the results.
%Towards this, we propose to use five late fusion approaches, as presented in Section~\ref{sec:exp_setup}.
For all models trained from scratch, we evaluated all 4 possible combinations of views for all 8 networks. Since we performed 5 different types of fusions and it was trained 3 different models from scratch, all the combinations would result in 184 experiments.
Given that high number of experiments and that, as as previously discussed, all networks produced similar results and any one could be selected for further experiments, we reported the obtained results for only the VGG~\cite{vgg}. 
%Those results can be seen in Figure~\ref{fig:resultsvgg-airound}. 
%However, we only reported, in Figure~\ref{fig:resultsvgg-airound}, the combination results for the VGG network~\cite{vgg}.
%This is because, as previously discussed, all networks produced similar results and any one could be selected for further experiments.
%Hence, in this case, we selected, for further consideration, the one with best average results, i.e., the VGG network~\cite{vgg}.

\begin{figure*}[!ht]%
    \centering
    \subfloat[Results comparison (in terms of balanced accuracy) of all fusion types using VGG trained from scratch.]{{\includegraphics[width=.97\columnwidth]{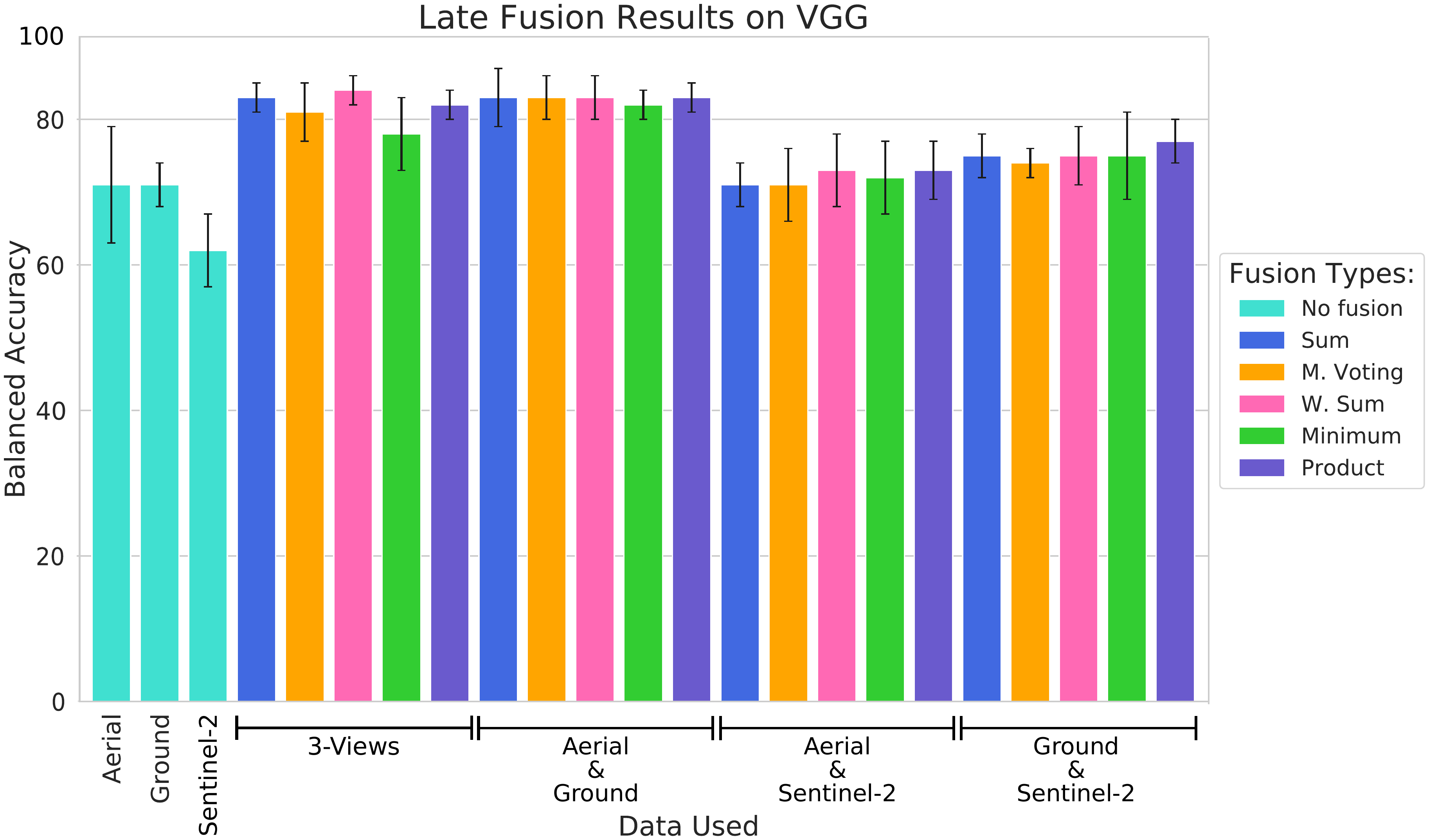} }}%
    \qquad
    \centering
    \subfloat[Results comparison (in terms of f1-score) of all fusion types using VGG trained from scratch.]{{\includegraphics[width=.97\columnwidth]{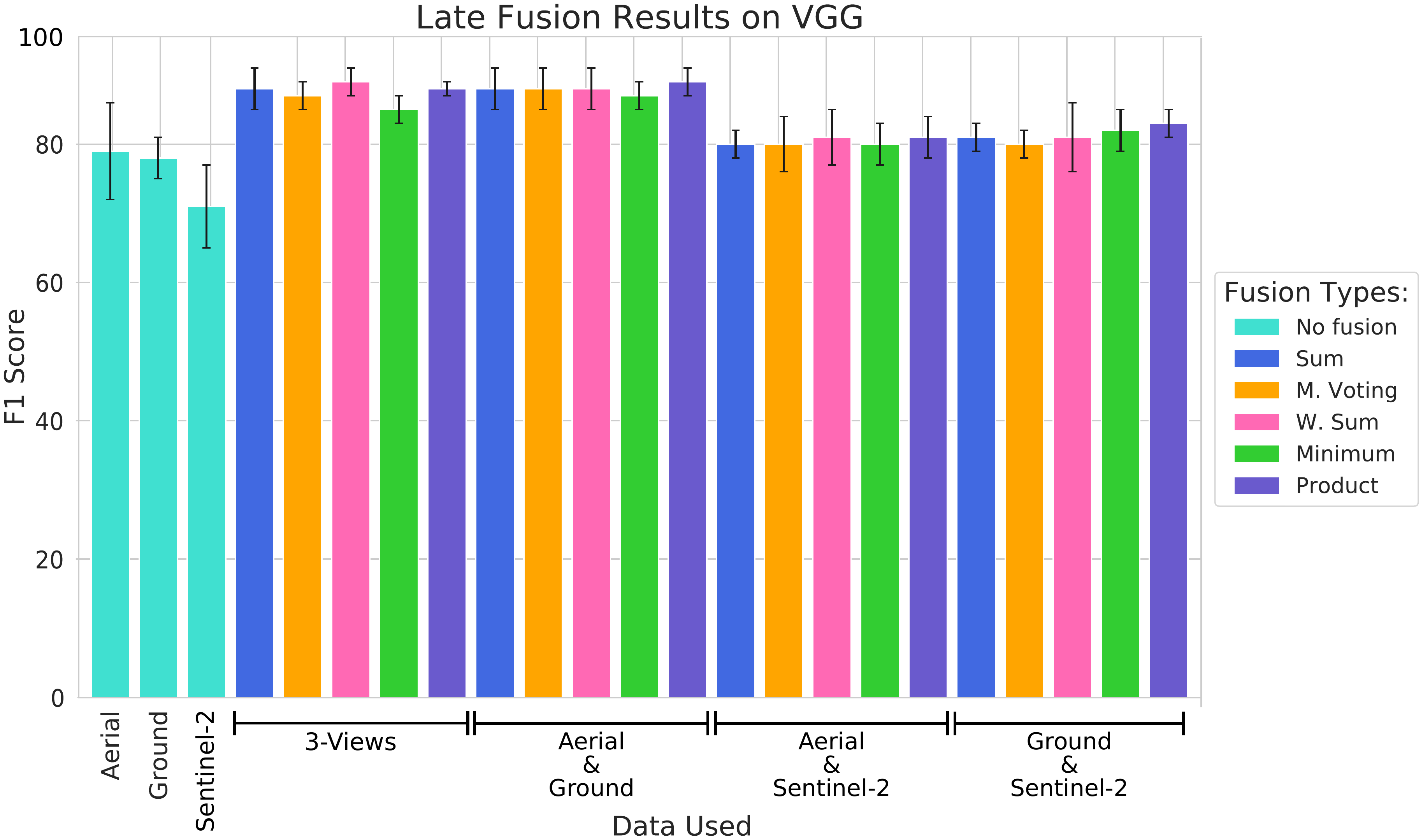}}}%
    \caption{Results comparison of all fusion types using VGG trained from scratch..}%
    \label{fig:resultsvgg-airound}%
\end{figure*}

As can be seen in Figure~\ref{fig:resultsvgg-airound}, most of the late fusion techniques outperformed the models trained with only one view, with a special highlight for the 3-view and aerial-ground fusions.
This result can be explained due to the amount of complementary information that exists between aerial and ground images.
On the other hand, based on all experiments, it is possible to observe that the combination of aerial and Sentinel-2 images tends to not statistically improve the results.
This is probably due to the amount of similar information that both types of images have in common, since both are from the same view perspective.

Finally, considering the ground-Sentinel-2 fusions, it is possible to notice a little improvement, that can be justified by the same reason of aerial-ground fusions.
We conclude that the gain in this fusion was not as good as aerial-ground one, because of the limited spatial-resolution that Sentinel-2 satellite offers ($10$\textit{m}$\times 10$\textit{m} or $20$\textit{m}$\times 20$\textit{m} or $60$\textit{m}$\times 60$\textit{m} per pixel, depending on the channel).

% \begin{figure}[]
% 	\centering
% 	   %\includegraphics[width=0.48\textwidth]{images/vgg_barplot.pdf}
% 	   \includegraphics[width=\columnwidth]{images/vgg_barplot.pdf}
% 	\caption{Results comparison of all fusion types using VGG trained from scratch.}
% 	\label{fig:resultsvgg-airound}
% \end{figure}

For the fine-tuned models, the late fusion results are presented in Table~\ref{tab:results-airound}.
In this case, all results are reported, given that Sentinel-2 images could not be exploited and only one combination could be performed (aerial and ground).

Comparing the results using only one type of data (Table~\ref{tab:baseline-airound}) with the fusion outcomes, it is possible to notice that the late fusion outperformed all approaches using only one view.
This corroborates with our initial analysis that the combination of multi-source data could improve the results for the scene classification task.
Furthermore, the late fusion results of the fine-tuned models yielded better results than the fusion results of the networks trained from scratch, an expected outcome.

\begin{table*}[!ht]
\centering
\begin{adjustbox}{max width=\textwidth}
\begin{tabular}{ccccccccccc}
\hline
\multirow{3}{*}{\textbf{Network}} & \multicolumn{10}{c}{\textbf{Fusion Strategy}} \\ \cline{2-11} 
 & \multicolumn{2}{c}{\textbf{Sum}} & \multicolumn{2}{c}{\textbf{M. Voting}} & \multicolumn{2}{c}{\textbf{W. Sum}} & \multicolumn{2}{c}{\textbf{Minimum}} & \multicolumn{2}{c}{\textbf{Product}} \\ \cline{2-11} 
 & \textbf{B. Acc.} & \textbf{F1 Score} & \textbf{B. Acc.} & \textbf{F1 Score} & \textbf{B. Acc.} & \textbf{F1 Score} & \textbf{B. Acc.} & \textbf{F1 Score} & \textbf{B. Acc.} & \textbf{F1 Score} \\ \hline
\textbf{AlexNet~\cite{alexnet}} & $0.87 \pm 0.06$ & $0.90 \pm 0.05$ & $0.86 \pm 0.08$ & $0.89 \pm 0.06$ & $0.87 \pm 0.08$ & $0.90 \pm 0.05$ & $0.87 \pm 0.05$ & $0.91 \pm 0.04$ & $0.89 \pm 0.06$ & $0.92 \pm 0.04$ \\
\textbf{VGG~\cite{vgg}} & $\mathbf{0.93 \pm 0.02}$ & $\mathbf{0.95 \pm 0.02}$ & $\mathbf{0.93 \pm 0.02}$ & $0.94 \pm 0.01$ & $0.92 \pm 0.02$ & $0.94 \pm 0.01$ & $0.92 \pm 0.06$ & $0.94 \pm 0.04$ & $0.93 \pm 0.04$ & $0.95 \pm 0.02$ \\
\textbf{Inception~\cite{inceptionv3}} & $\mathbf{0.93 \pm 0.03}$ & $\mathbf{0.95 \pm 0.02}$ & $\mathbf{0.93 \pm 0.03}$ & $\mathbf{0.95 \pm 0.02}$ & $\mathbf{0.93 \pm 0.04}$ & $\mathbf{0.95 \pm 0.02}$ & $\mathbf{0.93 \pm 0.03}$ & $\mathbf{0.95 \pm 0.01}$ & $\mathbf{0.94 \pm 0.03}$ & $0.95 \pm 0.01$ \\
\textbf{ResNet~\cite{resnet}} & $0.92 \pm 0.02$ & $0.94 \pm 0.01$ & $0.91 \pm 0.03$ & $0.93 \pm 0.02$ & $0.92 \pm 0.03$ & $0.94 \pm 0.01$ & $0.91 \pm 0.02$ & $0.93 \pm 0.01$ & $0.92 \pm 0.04$ & $0.94 \pm 0.02$ \\
\textbf{DenseNet~\cite{densenet}} & $\mathbf{0.93 \pm 0.03}$ & $\mathbf{0.95 \pm 0.02}$ & $0.92 \pm 0.03$ & $0.94 \pm 0.02$ & $0.92 \pm 0.03$ & $0.94 \pm 0.02$ & $0.92 \pm 0.03$ & $0.94 \pm 0.02$ & $0.93 \pm 0.04$ & $0.95 \pm 0.02$ \\
\textbf{SqueezeNet~\cite{squeezenet}} & $0.91 \pm 0.04$ & $0.93 \pm 0.03$ & $0.91 \pm 0.04$ & $0.92 \pm 0.03$ & $0.90 \pm 0.05$ & $0.92 \pm 0.02$ & $0.88 \pm 0.04$ & $0.91 \pm 0.02$ & $0.91 \pm 0.03$ & $0.93 \pm 0.02$ \\
\textbf{SENet~\cite{seresnet}} & $0.92 \pm 0.03$ & $0.93 \pm 0.02$ & $0.92 \pm 0.04$ & $0.93 \pm 0.02$ & $0.92 \pm 0.03$ & $0.93 \pm 0.02$ & $0.92 \pm 0.04$ & $0.94 \pm 0.02$ & $0.92 \pm 0.04$ & $0.94 \pm 0.02$ \\
\textbf{SKNet~\cite{sknet}} & $\mathbf{0.93 \pm 0.04}$ & $\mathbf{0.95 \pm 0.02}$ & $\mathbf{0.93 \pm 0.04}$ & $\mathbf{0.95 \pm 0.03}$ & $\mathbf{0.93 \pm 0.03}$ & $\mathbf{0.95 \pm 0.02}$ & $0.92 \pm 0.04$ & $\mathbf{0.95 \pm 0.02}$ & $\mathbf{0.94 \pm 0.04}$ & $\mathbf{0.96 \pm 0.02}$ \\ \bottomrule
\end{tabular}
\end{adjustbox}
\caption{Results of the evaluated late fusion techniques for \thedataset~dataset using fine-tuned models.}
\label{tab:results-airound}
\end{table*}

%%%%%%%%%%%%%%%%%%%%%%%%%%%%%%%%%%%%%%%%%%%%%%%%%%%%%%%%%%%%%%%%

In order to better understand how the fusion methods are able to improve the results, we performed an analysis, per class, of such techniques for all eight architectures.
However, again, because of the same aforementioned reason, only results for the VGG architecture~\cite{vgg} were reported.

%Figure~\ref{fig:vggimp-airound} reports the fusion results per class. 
%In this heat map, the values represent the ratio between the performance of a single VGG~\cite{vgg}, trained/fine-tuned in one specific domain, and a fusion of two VGGs~\cite{vgg}, trained/fine-tuned on the both aerial and ground views.
%Therefore, positive/blue values indicate that the classification of that class was improved when comparing the network trained on a specific view and the fusion method, while the negative/red values indicate that the classification of that class worsened.
%represent the gain that each class had after performing a fusion using the other domain's VGG~\cite{vgg} predictions.
%Specifically, the performance of a VGG~\cite{vgg}, trained in one specific domain, is compared to the outcome of the fusion of two VGGs~\cite{vgg} with the outputs generated after performing a fusion using the results of another VGG, trained in the other domain.

\begin{figure*}[!ht]
    \centering
    \subfloat[\thedataset~dataset.] {
        \includegraphics[width=.99\textwidth]{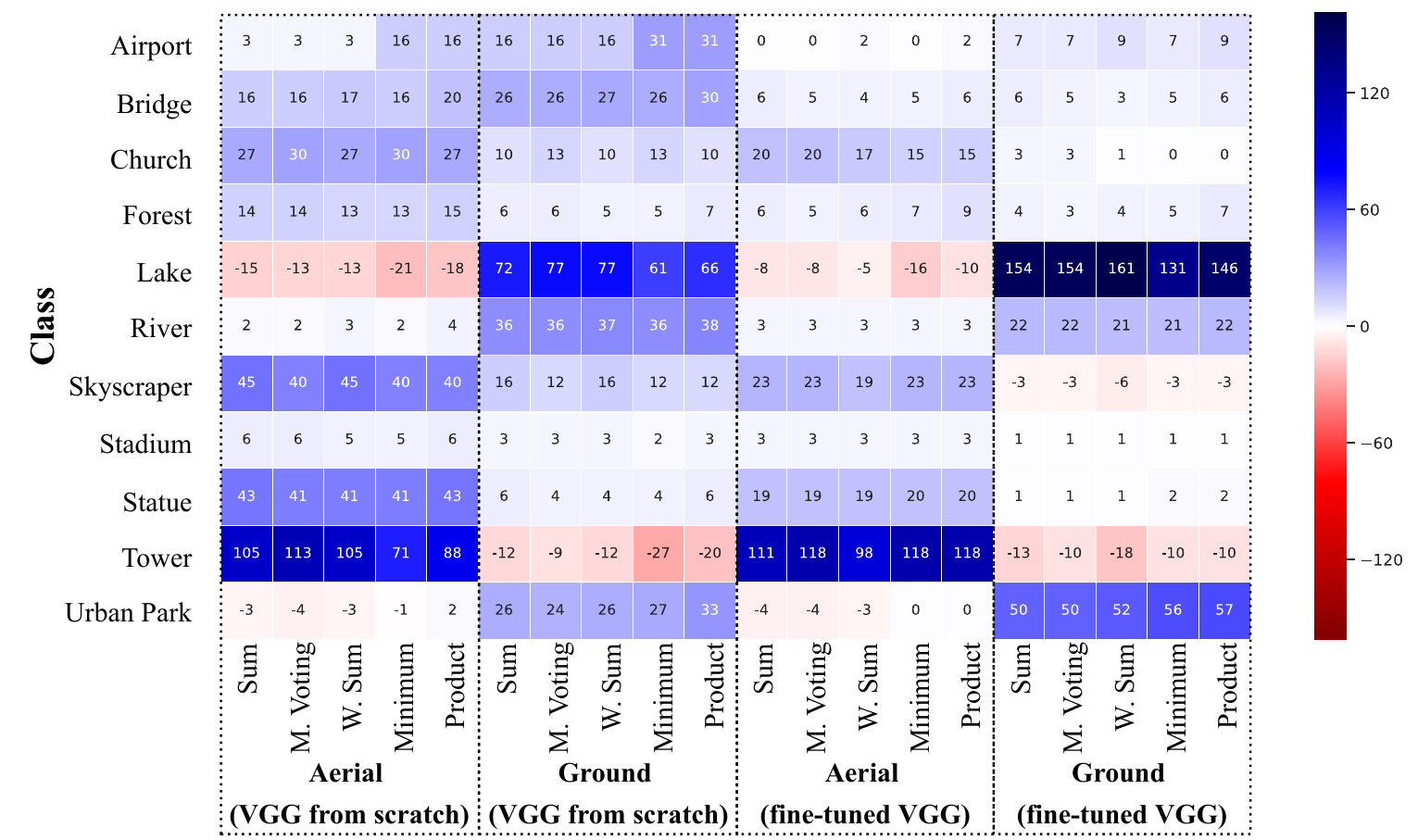}
        \label{fig:vggimp-airound}
    }
    \qquad
    \subfloat[CV-BrCT dataset.]{
        \includegraphics[width=.99\textwidth]{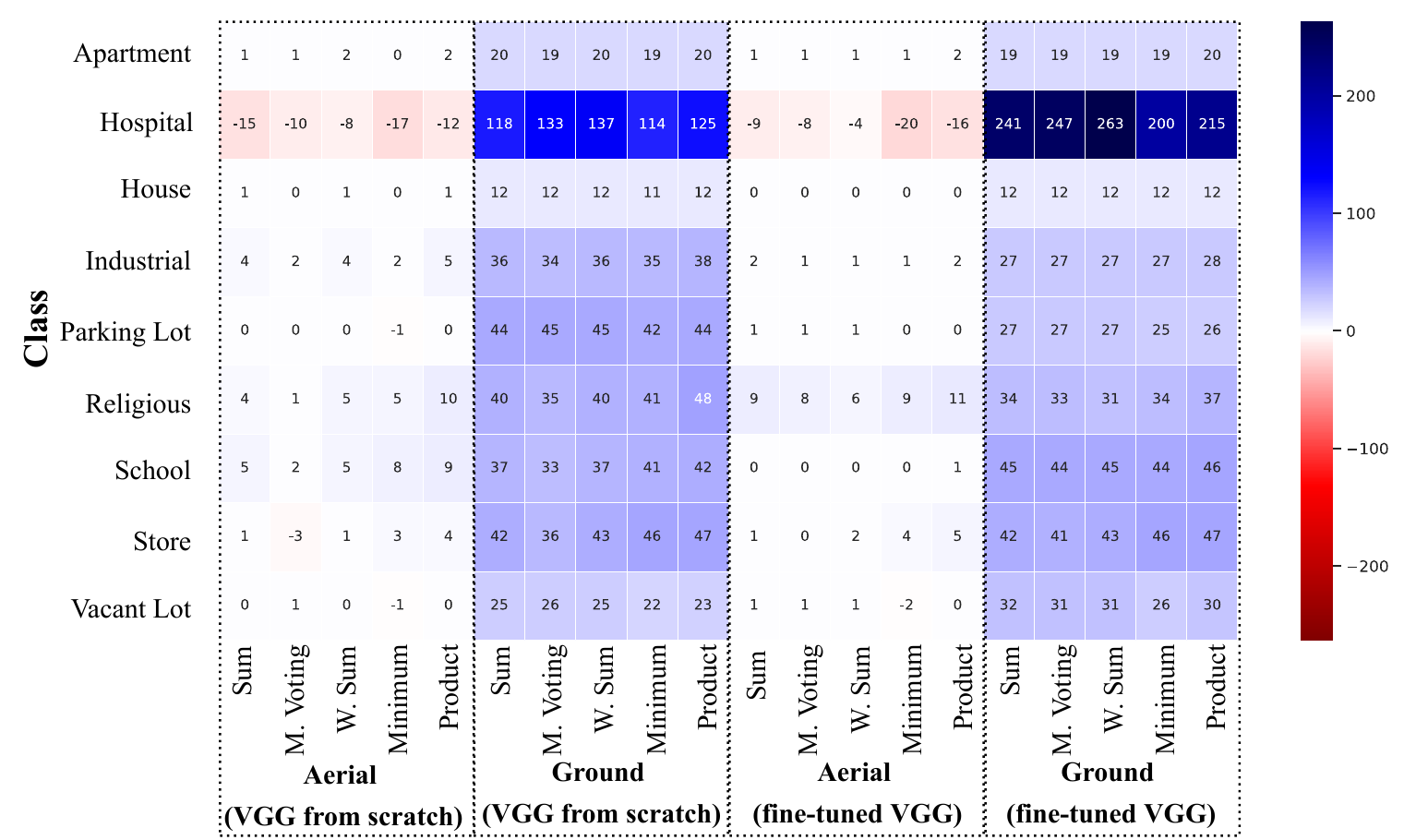} 
        \label{fig:vggimp-cvbrct}
    }
\caption{Values represent the ratio of the accuracy per class between a single VGG~\cite{vgg}, trained/fine-tuned in one specific domain, and a fusion of two VGGs~\cite{vgg}, trained/fine-tuned on both aerial and ground views.
In the numerator of this ratio, we calculated the difference between the accuracy pos-fusion and the accuracy using only one view, for each class.
Therefore, positive/blue values indicate that the classification of that class was improved when comparing the network trained on a specific view and the fusion method, while the negative/red values indicate that the classification of that class worsened.}
\end{figure*}

Figure~\ref{fig:vggimp-airound} reports the fusion improvements per class. As can be seen in the figure, this process was performed individually (per view and class) using models trained from scratch and fine-tuned.
This is because the purpose of this heat map is to see which classes benefit the most from each fusion in each type of view.

Through Figure~\ref{fig:vggimp-airound}, it is possible to observe that, for aerial data, the classes tower, skyscraper, and statue were the ones that most benefited from the aerial/ground fusion.
This is due to the fact that all of them are hard to classify using only aerial images since those structures naturally have high heights and occupy a restricted area, which are characteristics that are not well explored in an aerial perspective. Concerning the ground images, the classes lake, river, and urban park were the ones that most improved from the aerial/ground fusion.
The main reason for this is that the context around those classes may help a lot in discriminating them. Specifically, urban parks are naturally located in cities and, therefore, the information about the existence of a city nearby, that come from aerial images, may help in its classification.
Furthermore, classes river and lake are quite similar, since both represent water bodies.
Thus, both classes may benefit from the information that aerial data provides about the area (such as the surrounding vegetation) which may help in these two classes' discrimination.

% \begin{figure*}[!ht]
% 	\centering
% % 	\includegraphics[width=.96\textwidth]{images/vgg-imp.pdf}
% 	\includegraphics[width=.90\textwidth]{images/heatmap_multiview.pdf}
% 	\caption{Values represent the ratio of the accuracy per class between a single VGG~\cite{vgg}, trained/fine-tuned in one specific domain, and a fusion of two VGGs~\cite{vgg}, trained/fine-tuned on both aerial and ground views.
%     In the numerator of this ratio, we calculated the difference between the accuracy pos-fusion and the accuracy using only one view, for each class.
%     Therefore, positive/blue values indicate that the classification of that class was improved when comparing the network trained on a specific view and the fusion method, while the negative/red values indicate that the classification of that class worsened.}
% 	\label{fig:vggimp-airound}
% \end{figure*}

\noindent \textbf{Remark}. Comparing the results between early and late fusion we can see a clear advantage of late fusion in most of the cases. 
However, in some situations, early fusion achieved competitive results, as can be seen comparing the results for fine-tuned DenseNets~\cite{densenet}, for instance.
Lastly, 
%since we stated that one of the great advantages of early fusion models comparing to late fusion ones is related to the sizes of the models and the necessary time to train them.  
we evaluated a series of experiments, measuring the spent time to train each one of the models in both datasets. Table~\ref{tab:times} contain those results, demonstrating clearly the aforementioned advantages of early fusion for \thedataset~dataset.

\begin{table}[!ht]
\centering
\begin{adjustbox}{max width=\columnwidth}
\begin{tabular}{@{}ccccc@{}}
\toprule
\textbf{Fusion Type} & \textbf{Network} & \textbf{\begin{tabular}[c]{@{}c@{}}GPU Train Time \\ AiRound\\ (in seconds)\end{tabular}} & \textbf{\begin{tabular}[c]{@{}c@{}}GPU Train Time \\ CV-BrCT\\ (in seconds)\end{tabular}} & \textbf{\begin{tabular}[c]{@{}c@{}}Total \\ Parameters\\ (in millions)\end{tabular}} \\ \midrule
\multirow{8}{*}{\textbf{Early Fusion}}
 & \textbf{AlexNet~\cite{alexnet}} & 34.98 & 294.00 & 56.72 \\
 & \textbf{VGG~\cite{vgg}} & 41.95 & 600.00 & 128.60 \\
 & \textbf{Inception~\cite{inceptionv3}} & 636.92 & 5226.10 & 24.37 \\
 & \textbf{ResNet~\cite{resnet}} & 59.16 & 736.40 & 10.81 \\
 & \textbf{DenseNet~\cite{densenet}} & 527.04 & 6215.60 & 12.42 \\
 & \textbf{SqueezeNet~\cite{squeezenet}} & 89.47 & 1107.60 & 0.73 \\
 & \textbf{SENet~\cite{seresnet}} & 461.42 & 5276.00 & 24.90 \\
 & \textbf{SKNet~\cite{sknet}} & 1705.60 & 20856.60 & 42.60 \\ \midrule
\multirow{8}{*}{\textbf{Late Fusion}}
 & \textbf{AlexNet~\cite{alexnet}} & 129.92 & 361.95 & 114.09 \\
 & \textbf{VGG~\cite{vgg}} & 144.54 & 994.24 & 268.62 \\
 & \textbf{Inception~\cite{inceptionv3}} & 1231.75 & 10284.88 & 48.84 \\
 & \textbf{ResNet~\cite{resnet}} & 207.37 & 1164.06 & 22.36 \\
 & \textbf{DenseNet~\cite{densenet}} & 984.54 & 11197.90 & 27.20 \\
 & \textbf{SqueezeNet~\cite{squeezenet}} & 264.29 & 1604.75 & 1.48 \\
 & \textbf{SENet~\cite{seresnet}} & 1166.22 & 9511.95 & 52.08 \\
 & \textbf{SKNet~\cite{sknet}} & 4942.72 & 33189.20 & 87.30 \\ \bottomrule
\end{tabular}
\end{adjustbox}
\caption{Benchmarched methods properties. It is important to mention that all the times were calculated using a RTX2080TI and it was accounted only forward and backward time during the training phase.}
\label{tab:times}
\end{table}

%%%%%%%%%%%%%%%%%%%%%%%%%%%%%%%%%%%%%%%%%%%%%%%%%%%%%%%%%%%%%%%%%%%%%%%%%%%%%%%%%%%%%%%%%%%%%%%%%%%%%%%%%%%%%%%%%%%%%%%%%%%%%%%%%%%%%%%%%%%%%%%%%%%%%%%%%%%%%%%%%%%%%%%%%%%%%%%%%%%%%%%%%%%%%%%%%%%%%%%%%%%%%%%%%%

\subsection{Experiments on the CV-BrCT} \label{sec:result_cv}

%\todo{Edemir e Pedro, escrever essa parte.}

\subsubsection{Networks Architectures Comparison} \label{sec:networks_cv}
We replicate the same experiments realized in the \thedataset~dataset in the CV-BrCT dataset in regards to its two image types - aerial and ground (frontal). The results for the experiments using a single type of image are presented in the table~\ref{tab:results-CV-BrCT}.

As seen in the table~\ref{tab:results-CV-BrCT} the best training protocol is again to fine-tune the networks. For all architectures, the fine-tuned models have better performance for both types of images, a result similar to the \thedataset~dataset experiments.  
Different from the other experiment, in the CV-BrCT dataset the networks tend to perform better with the aerial images than with the ground images, while in the \thedataset~dataset these results were closer. 
We argue that some classes have a visual similarity in the ground images, e.g. hospitals and schools can have a similar facades, prevalence of stores in first floors of buildings, etc. Thus the discrepancy between results of different image types.

With a few exceptions, the networks have comparable results and four achieved practically the same metric values.

\begin{table}[!ht]
\setlength{\tabcolsep}{5pt}
\begin{adjustbox}{max width=\columnwidth}
\begin{tabular}{@{}ccllll@{}}
\toprule
\multicolumn{1}{l}{\multirow{4}{*}{\textbf{\begin{tabular}[c]{@{}l@{}}Training \\ Strategy\end{tabular}}}} & \multirow{4}{*}{\textbf{Network}} & \multicolumn{4}{c}{\textbf{Input Data}} \\ \cmidrule(l){3-6} 
\multicolumn{1}{l}{} &  & \multicolumn{2}{c}{\textbf{Aerial}} & \multicolumn{2}{c}{\textbf{Ground}} \\ \cmidrule(l){3-6} 
\multicolumn{1}{l}{} &  & \textbf{B. Acc.} & \textbf{F1 Score} & \textbf{B. Acc.} & \textbf{F1 Score} \\ \midrule
\multirow{8}{*}{\textbf{\begin{tabular}[c]{@{}c@{}}Training\\ from\\ scratch\end{tabular}}}
 & \textbf{AlexNet~\cite{alexnet}} & $0.68 \pm 0.03$ & $0.79 \pm 0.02$ & $0.50 \pm 0.03$ & $0.62 \pm 0.01$ \\
 & \textbf{VGG~\cite{vgg}} & $0.70 \pm 0.04$ & $\mathbf{0.81 \pm 0.03}$ & $\mathbf{0.54 \pm 0.02}$ & $\mathbf{0.66 \pm 0.01}$ \\
 & \textbf{Inception~\cite{inceptionv3}} & $0.69 \pm 0.03$ & $0.80 \pm 0.02$ & $0.49 \pm 0.03$ & $0.62 \pm 0.02$ \\
 & \textbf{ResNet~\cite{resnet}} & $0.68 \pm 0.07$ & $0.79 \pm 0.03$ & $0.50 \pm 0.05$ & $0.63 \pm 0.03$ \\
 & \textbf{DenseNet~\cite{densenet}} & $\mathbf{0.71 \pm 0.02}$ & $\mathbf{0.81 \pm 0.01}$ & $0.49 \pm 0.01$ & $0.62 \pm 0.01$ \\
 & \textbf{SqueezeNet~\cite{squeezenet}} & $0.55 \pm 0.07$ & $0.70 \pm 0.05$ & $0.41 \pm 0.08$ & $0.56 \pm 0.06$ \\
 & \textbf{SENet~\cite{seresnet}} & $0.69 \pm 0.04$ & $0.80 \pm 0.02$ & $0.49 \pm 0.02$ & $0.62 \pm 0.02$ \\
 & \textbf{SKNet~\cite{sknet}} & $0.68 \pm 0.06$ & $0.79 \pm 0.04$ & $0.47 \pm 0.03$ & $0.61 \pm 0.02$ \\ \midrule
\multirow{8}{*}{\textbf{\begin{tabular}[c]{@{}c@{}}Fine Tuning\\ from\\ ImageNet\end{tabular}}}
 & \textbf{AlexNet~\cite{alexnet}} & $0.75 \pm 0.02$ & $0.84 \pm 0.01$ & $0.54 \pm 0.01$ & $0.66 \pm 0.01$ \\
 & \textbf{VGG~\cite{vgg}} & $0.79 \pm 0.03$ & $0.87 \pm 0.01$ & $\mathbf{0.60 \pm 0.02}$ & $\mathbf{0.71 \pm 0.01}$ \\
 & \textbf{Inception~\cite{inceptionv3}} & $\mathbf{0.80 \pm 0.02}$ & $0.87 \pm 0.00$ & $\mathbf{0.60 \pm 0.03}$ & $\mathbf{0.71 \pm 0.01}$ \\
 & \textbf{ResNet~\cite{resnet}} & $0.78 \pm 0.02$ & $0.86 \pm 0.01$ & $0.58 \pm 0.04$ & $0.69 \pm 0.02$ \\
 & \textbf{DenseNet~\cite{densenet}} & $\mathbf{0.80 \pm 0.02}$ & $0.87 \pm 0.01$ & $\mathbf{0.60 \pm 0.01}$ & $\mathbf{0.71 \pm 0.01}$ \\
 & \textbf{SqueezeNet~\cite{squeezenet}} & $0.70 \pm 0.02$ & $0.80 \pm 0.01$ & $0.56 \pm 0.02$ & $0.68 \pm 0.01$ \\
 & \textbf{SENet~\cite{seresnet}} & $\mathbf{0.80 \pm 0.02}$ & $0.87 \pm 0.01$ & $\mathbf{0.60 \pm 0.01}$ & $\mathbf{0.71 \pm 0.01}$ \\
 & \textbf{SKNet~\cite{sknet}} & $\mathbf{0.80 \pm 0.03}$ & $\mathbf{0.88 \pm 0.01}$ & $\mathbf{0.60 \pm 0.02}$ & $\mathbf{0.71 \pm 0.01}$ \\ \bottomrule
\end{tabular}
\end{adjustbox}
\caption{Results of the evaluated models for CV-BrCT dataset.}
\label{tab:results-CV-BrCT}
\end{table}

\subsubsection{Fusions} \label{sec:fusion_cv}

In this next section we present the results for the fusion methods in the CV-BrCT dataset. 
%Table~\ref{tab:times} also contains the spent time to train the models in CV-BrCT dataset, and again it demonstrates that early fusion models tend to converge faster than late fusion ones. 

\noindent \textbf{CV-BrCT Early Fusion}. The early fusion architectures proposed were evaluated with pretrained weights and initially randomized weights. The results are presented in the table~\ref{tab:results-CV-BrCT-early}. As in the experiments, the fine-tuned models outperform the non-pretrained ones. In respect to the single type networks, these early fusion architectures seem to perform slightly better than the trained from scratch with one type of image, while performing the same, or slightly worse, than the fine tuned models using aerial images. 

% As expected, once the results of the networks using only ground images were worse than the aerial models, letting the networks merge and combine the features of both images, in the fine tunning scenario, lead to no improvements.
As noted in the previous experiment, the results of networks using only ground images were worse than the aerial image models. 
Through the experiments, we noted that using a network that merges and combines features of both images from the start leads to no improvements, in the fine-tune scenario. This issue can also be justified by the same feature degradation phenomenon, aforementioned.
However, when the networks are trained from scratch it seems to learn how to better extract and combine features of both images, which yields a slightly superior performance of these models, comparing to the results previously reported in Table~\ref{tab:results-CV-BrCT}.

\begin{table}[!ht]
\centering
\begin{adjustbox}{max width=\columnwidth}
\begin{tabular}{@{}ccccc@{}}
\toprule
\multirow{4}{*}{\textbf{\begin{tabular}[c]{@{}c@{}}Early Fusion\\ Networks\end{tabular}}} & \multicolumn{4}{c}{\textbf{Training Strategy}} \\ \cmidrule(l){2-5} 
 & \multicolumn{2}{c}{\textbf{From Scratch}} & \multicolumn{2}{c}{\textbf{Fine Tuning}} \\ \cmidrule(l){2-5} 
 & \textbf{B. Acc.} & \textbf{F1 Score} & \textbf{B. Acc.} & \textbf{F1 Score} \\ \midrule
\textbf{AlexNet~\cite{alexnet}} & $0.69 \pm 0.03$ & $0.8 \pm 0.01$ & $0.72 \pm 0.02$ & $0.82 \pm 0.01$ \\
\textbf{VGG~\cite{vgg}} & $\mathbf{0.73 \pm 0.03}$ & $0.82 \pm 0.01$ & $0.76 \pm 0.02$ & $0.84 \pm 0.02$ \\
\textbf{Inception~\cite{inceptionv3}} & $\mathbf{0.73 \pm 0.04}$ & $\mathbf{0.83 \pm 0.02}$ & $0.79 \pm 0.03$ & $\mathbf{0.87 \pm 0.01}$ \\
\textbf{ResNet~\cite{resnet}} & $0.68 \pm 0.02$ & $0.79 \pm 0.01$ & $0.74 \pm 0.02$ & $0.83 \pm 0.01$ \\
\textbf{DenseNet~\cite{densenet}} & $0.71 \pm 0.04$ & $0.80 \pm 0.02$ & $0.72 \pm 0.03$ & $0.81 \pm 0.01$ \\
\textbf{SqueezeNet~\cite{squeezenet}} & $0.60 \pm 0.01$ & $0.73 \pm 0.02$ & $0.67 \pm 0.04$ & $0.79 \pm 0.02$ \\
\textbf{SENet~\cite{seresnet}} & $0.67 \pm 0.04$ & $0.78 \pm 0.02$ & $0.78 \pm 0.02$ & $0.86 \pm 0.01$ \\
\textbf{SKNet~\cite{sknet}} & $0.70 \pm 0.04$ & $0.80 \pm 0.03$ & $\mathbf{0.80 \pm 0.02}$ & $\mathbf{0.87 \pm 0.01}$ \\ \bottomrule
\end{tabular}
\end{adjustbox}
\caption{Results of the evaluated early fusion networks for CV-BrCT dataset.}
\label{tab:results-CV-BrCT-early}
\end{table}

\noindent \textbf{CV-BrCT Late Fusion}. We tested the fusion of the two image types in all the five methods discussed in section~\ref{sec:late-fusion-methods}. All the results are show in table~\ref{tab:results-CV-BrCT-ft}.

Overall, all fusion methods improved the results of the networks trained with a single type, in both the initially randomized and fine-tuned cases. The results across fusion methods are similar, although some techniques show a consistent improvement, e.g., Weighted Sum, and other do not appears to have a noticeable effect, e.g., Minimum. 

As the networks trained with only ground images are less reliable classifiers, i.e., have achieved worse results than the aerial models, the score each one assign to a sample is smaller than the aerial model. Henceforth, the impact these classifiers have in the final prediction, regardless of the fusion method, are less significant, thus the improvement exists but are relatively small.

Similar to Figure~\ref{fig:vggimp-airound}, we also produced a figure to the CV-BrCT dataset. In Figure~\ref{fig:vggimp-cvbrct}, we can see the impact of the different fusion methods in each class of the dataset, for each single image type network model (in this case the VGG model). 
As we can see, all classes have an improvement in relation to the single type networks of ground images. Furthermore, the Hospital class is the one mostly impacted by the addition of the aerial data. 
As hospital, usually, have large footprints an single image from a frontal perspective can capture a facade easily confoundable with other classes facades. Consequently, the addition of an aerial view can distinguish an ambiguous hospital sample. 
In contrast, the aerial models display few improvements - probably a few ambiguous samples were corrected by frontal images - to all classes but Hospitals.

\noindent \textbf{Remark}. As can be noted in Tables \ref{tab:results-CV-BrCT-early}, and~\ref{tab:results-CV-BrCT-ft}, for CV-BrCT, the late fusion models tended to achieve slightly better results than the early fusion networks. 
Again, some early fusion models achieved competitive results compared to the late fusion ones.
For instance, analyzing models trained from scratch, the early fusion adaptation of Inception~\cite{inceptionv3} matches to the late fusion approach for the same network, and for fine-tuned models, the same happened to Selective Kernels Network~\cite{sknet}.
Lastly, we would like to highlight the times spent to train each model in the CV-BrCT dataset. As can be noted in Table~\ref{tab:times}, it also demonstrates that early fusion models tend to converge faster than late fusion ones.

%\todo{Lembrar de comentar da figura.}

% \begin{figure*}[!ht]
% 	\centering
% % 	\includegraphics[width=.96\textwidth]{images/vgg-imp.pdf}
% 	\includegraphics[width=.90\textwidth]{images/heatmap_cvbrct.pdf}
% 	\caption{This heatmap was designed following the same scheme used in Figure \ref{fig:vggimp-airound}, but instead of using \thedataset~results, it uses CV-BrCT.}
% 	\label{fig:vggimp-cvbrct}
% \end{figure*}

\begin{table*}[!ht]
\centering
\begin{adjustbox}{max width=\textwidth}
\begin{tabular}{cccccccccccc}
\toprule
\multirow{4}{*}{\textbf{\begin{tabular}[c]{@{}c@{}}Training\\ Strategy\end{tabular}}} & \multirow{4}{*}{\textbf{Network}} & \multicolumn{10}{c}{\textbf{Fusion Strategy}} \\ 
\cmidrule{3-12} 
&  & \multicolumn{2}{c}{\textbf{Sum}} & \multicolumn{2}{c}{\textbf{M. Voting}} & \multicolumn{2}{c}{\textbf{W. Sum}} & \multicolumn{2}{c}{\textbf{Minimum}} & \multicolumn{2}{c}{\textbf{Product}} \\ 
 \cmidrule{3-12} 
&  & \textbf{B. Acc.} & \textbf{F1 Score} & \textbf{B. Acc.} & \textbf{F1 Score} & \textbf{B. Acc.} & \textbf{F1 Score} & \textbf{B. Acc.} & \textbf{F1 Score} & \textbf{B. Acc.} & \textbf{F1 Score} \\ 
 \midrule
\multirow{8}{*}{\textbf{\begin{tabular}[c]{@{}c@{}}Training\\ from\\ Scratch\end{tabular}}} & \textbf{AlexNet~\cite{alexnet}} & $0.69 \pm 0.03$ & $0.80 \pm 0.02$ & $0.68 \pm 0.03$ & $0.80 \pm 0.02$ & $0.69 \pm 0.03$ & $0.81 \pm 0.02$ & $0.68 \pm 0.03$ & $0.80 \pm 0.02$ & $0.70 \pm 0.03$ & $0.82 \pm 0.01$ \\
& \textbf{VGG~\cite{vgg}} & $\mathbf{0.72 \pm 0.03}$ & $\mathbf{0.82 \pm 0.02}$ & $\mathbf{0.71 \pm 0.03}$ & $\mathbf{0.81 \pm 0.02}$ & $\mathbf{0.72 \pm 0.03}$ & $0.82 \pm 0.02$ & $0.71 \pm 0.02$ & $\mathbf{0.82 \pm 0.01}$ & $0.72 \pm 0.02$ & $\mathbf{0.83 \pm 0.02}$ \\
& \textbf{Inception~\cite{inceptionv3}} & $0.69 \pm 0.02$ & $0.81 \pm 0.02$ & $0.69 \pm 0.03$ & $0.80 \pm 0.02$ & $0.70 \pm 0.02$ & $0.81 \pm 0.02$ & $0.69 \pm 0.03$ & $0.81 \pm 0.02$ & $0.70 \pm 0.02$ & $0.82 \pm 0.02$ \\
& \textbf{ResNet~\cite{resnet}} & $0.69 \pm 0.06$ & $0.81 \pm 0.03$ & $0.68 \pm 0.06$ & $0.80 \pm 0.03$ & $0.70 \pm 0.06$ & $0.81 \pm 0.03$ & $0.68 \pm 0.06$ & $0.81 \pm 0.03$ & $0.70 \pm 0.06$ & $0.82 \pm 0.03$ \\
& \textbf{DenseNet~\cite{densenet}} & $\mathbf{0.72 \pm 0.03}$ & $\mathbf{0.82 \pm 0.02}$ & $\mathbf{0.71 \pm 0.02}$ & $\mathbf{0.81 \pm 0.02}$ & $\mathbf{0.72 \pm 0.02}$ & $\mathbf{0.83 \pm 0.02}$ & $\mathbf{0.71 \pm 0.02}$ & $\mathbf{0.82 \pm 0.02}$ & $\mathbf{0.73 \pm 0.03}$ & $\mathbf{0.83 \pm 0.02}$ \\
& \textbf{SqueezeNet~\cite{squeezenet}} & $0.57 \pm 0.05$ & $0.72 \pm 0.04$ & $0.56 \pm 0.05$ & $0.70 \pm 0.04$ & $0.57 \pm 0.05$ & $0.72 \pm 0.04$ & $0.56 \pm 0.04$ & $0.72 \pm 0.03$ & $0.57 \pm 0.05$ & $0.73 \pm 0.04$ \\
& \textbf{SENet~\cite{seresnet}} & $0.70 \pm 0.04$ & $0.81 \pm 0.02$ & $0.69 \pm 0.04$ & $0.80 \pm 0.02$ & $0.70 \pm 0.04$ & $0.81 \pm 0.02$ & $0.69 \pm 0.03$ & $0.80 \pm 0.02$ & $0.70 \pm 0.04$ & $0.82 \pm 0.02$ \\
& \textbf{SKNet~\cite{sknet}} & $0.69 \pm 0.06$ & $0.80 \pm 0.04$ & $0.68 \pm 0.06$ & $0.79 \pm 0.04$ & $0.69 \pm 0.05$ & $0.80 \pm 0.03$ & $0.67 \pm 0.04$ & $0.79 \pm 0.03$ & $0.69 \pm 0.05$ & $0.81 \pm 0.03$ \\ 
% \bottomrule
% \end{tabular}
% \end{adjustbox}
% \caption{Results of the evaluated late fusion techniques for CV-BrCT dataset training each one of the models from scratch.}
% \label{tab:results-CV-BrCT-fusion}
% \end{table*}

% \begin{table*}[]
% \centering
% %\setlength{\tabcolsep}{5pt}
% \begin{adjustbox}{max width=\textwidth}
% \begin{tabular}{ccccccccccc}
% \toprule
% \multirow{3}{*}{\textbf{Network}} & \multicolumn{10}{c}{\textbf{Fusion Strategy}} \\ \cmidrule{2-11} 
%  & \multicolumn{2}{c}{\textbf{Sum}} & \multicolumn{2}{c}{\textbf{M. Voting}} & \multicolumn{2}{c}{\textbf{W. Sum}} & \multicolumn{2}{c}{\textbf{Minimum}} & \multicolumn{2}{c}{\textbf{Product}} \\ \cmidrule{2-11} 
%  & \textbf{B. Acc.} & \textbf{F1 Score} & \textbf{B. Acc.} & \textbf{F1 Score} & \textbf{B. Acc.} & \textbf{F1 Score} & \textbf{B. Acc.} & \textbf{F1 Score} & \textbf{B. Acc.} & \textbf{F1 Score} \\ 
\midrule
\multirow{8}{*}{\textbf{\begin{tabular}[c]{@{}c@{}}Fine \\ Tuning\end{tabular}}} & \textbf{AlexNet~\cite{alexnet}} & $0.76 \pm 0.03$ & $0.85 \pm 0.02$ & $0.75 \pm 0.03$ & $0.84 \pm 0.02$ & $0.76 \pm 0.03$ & $0.85 \pm 0.01$ & $0.75 \pm 0.02$ & $0.85 \pm 0.01$ & $0.76 \pm 0.02$ & $0.86 \pm 0.01$ \\
& \textbf{VGG~\cite{vgg}} & $0.80 \pm 0.03$ & $0.88 \pm 0.01$ & $0.80 \pm 0.03$ & $\mathbf{0.88 \pm 0.01}$ & $0.80 \pm 0.03$ & $0.88 \pm 0.01$ & $0.79 \pm 0.02$ & $\mathbf{0.88 \pm 0.01}$ & $0.80 \pm 0.02$ & $0.88 \pm 0.01$ \\
& \textbf{Inception~\cite{inceptionv3}} & $\mathbf{0.81 \pm 0.01}$ & $0.88 \pm 0.01$ & $0.80 \pm 0.02$ & $\mathbf{0.88 \pm 0.01}$ & $\mathbf{0.81 \pm 0.02}$ & $0.88 \pm 0.00$ & $\mathbf{0.80 \pm 0.02}$ & $\mathbf{0.88 \pm 0.01}$ & $\mathbf{0.81 \pm 0.01}$ & $0.89 \pm 0.01$ \\
& \textbf{ResNet~\cite{resnet}} & $0.78 \pm 0.02$ & $0.87 \pm 0.01$ & $0.78 \pm 0.02$ & $0.86 \pm 0.01$ & $0.78 \pm 0.02$ & $0.87 \pm 0.01$ & $0.77 \pm 0.03$ & $0.87 \pm 0.01$ & $0.79 \pm 0.02$ & $0.87 \pm 0.01$ \\
& \textbf{DenseNet~\cite{densenet}} & $0.81 \pm 0.02$ & $0.88 \pm 0.01$ & $0.80 \pm 0.02$ & $\mathbf{0.88 \pm 0.01}$ & $\mathbf{0.81 \pm 0.03}$ & $0.88 \pm 0.01$ & $\mathbf{0.80 \pm 0.02}$ & $\mathbf{0.88 \pm 0.01}$ & $\mathbf{0.81 \pm 0.02}$ & $0.89 \pm 0.01$ \\
& \textbf{SqueezeNet~\cite{squeezenet}} & $0.72 \pm 0.02$ & $0.83 \pm 0.01$ & $0.72 \pm 0.02$ & $0.82 \pm 0.00$ & $0.73 \pm 0.02$ & $0.83 \pm 0.00$ & $0.71 \pm 0.02$ & $0.83 \pm 0.01$ & $0.73 \pm 0.02$ & $0.84 \pm 0.01$ \\
& \textbf{SENet~\cite{seresnet}} & $\mathbf{0.81 \pm 0.02}$ & $0.88 \pm 0.00$ & $\mathbf{0.81 \pm 0.02}$ & $\mathbf{0.88 \pm 0.00}$ & $\mathbf{0.81 \pm 0.02}$ & $0.88 \pm 0.00$ & $\mathbf{0.80 \pm 0.02}$ & $\mathbf{0.88 \pm 0.01}$ & $\mathbf{0.81 \pm 0.02}$ & $0.89 \pm 0.01$ \\
& \textbf{SKNet~\cite{sknet}} & $\mathbf{0.81 \pm 0.04}$ & $\mathbf{0.89 \pm 0.01}$ & $\mathbf{0.81 \pm 0.04}$ & $\mathbf{0.88 \pm 0.01}$ & $\mathbf{0.81 \pm 0.03}$ & $\mathbf{0.89 \pm 0.01}$ & $\mathbf{0.80 \pm 0.04}$ & $\mathbf{0.88 \pm 0.02}$ & $\mathbf{0.81 \pm 0.04}$ & $\mathbf{0.89 \pm 0.02}$ \\ 
\bottomrule
\end{tabular}
\end{adjustbox}
\caption{Results of the evaluated late fusion techniques for CV-BrCT dataset.}
\label{tab:results-CV-BrCT-ft}
\end{table*}

\section{Conclusion}
\label{sec:conclusion}
%\todo{ARRUMAR CONCLUSAO DE ACORDO COM O NOVO JEITO QUE O PAPER FOI FEITO.}
In this work we introduced two new publicly available datasets for multi-view image tasks, which were named \thedataset~and CV-BrCT. 
We conducted extensive experiments in which results can be summarized as:
(1) early and late fusion-based aerial + ground feature combination yielded very relevant results, but there is still room for improvements, specially in CV-BrCT dataset;
(2) fine-tuned models with feature fusion are quite effective;
(3) some classes in the dataset were unable to benefit from the multispectral information present in the sentinel-2 images from \thedataset.

%As it was shown by the benchmark made, this dataset can be used to evaluate image classification, feature fusion, and unsupervised domain adaptation techniques. Besides that,  algorithms for other tasks such as cross-view matching can also be benchmarked using \thedataset.

As future work, we foresee to expand both datasets, adding more classes and instances to the existing ones.
We also point out the need for deeper studies about cross-view matching algorithms, multi-view domain adaptation and more sophisticated feature fusion techniques, such as hybrid fusion or techniques that can handle well with lack of data or the presence of noisy images.

%\clearpage
%\pagebreak
\bibliographystyle{IEEEtran}
\bibliography{references.bib}

\end{document}